\documentclass[10pt,twocolumn,letterpaper]{article}

\usepackage{cvpr}              

\usepackage{graphicx}
\usepackage{amsmath}
\usepackage{amsthm}
\usepackage{amssymb}
\usepackage{booktabs}

\usepackage[ruled]{algorithm2e}
\usepackage{bbm}
\usepackage{bm}
\usepackage{color}
\usepackage{multirow}
\usepackage{array}
\usepackage{booktabs}
\usepackage[switch]{lineno}

\newcommand{\Tref}[1]{Table~\ref{#1}}

\newcommand{\Fref}[1]{Figure~\ref{#1}}

\usepackage[pagebackref,breaklinks,colorlinks,linkcolor=blue]{hyperref}

\usepackage[accsupp]{axessibility}  

\usepackage[capitalize]{cleveref}
\crefname{section}{Sec.}{Secs.}
\Crefname{section}{Section}{Sections}
\Crefname{table}{Table}{Tables}
\crefname{table}{Tab.}{Tabs.}

\def\cvprPaperID{5230} 
\def\confName{CVPR}
\def\confYear{2022}

\begin{document}

\title{Shape-invariant 3D Adversarial Point Clouds}

\author{
Qidong Huang\textsuperscript{\rm 1} \qquad
Xiaoyi Dong\textsuperscript{\rm 1} \qquad
Dongdong Chen\textsuperscript{\rm 2} \qquad
Hang Zhou\textsuperscript{\rm 3} \\
Weiming Zhang\textsuperscript{\rm 1,}\thanks{Corresponding author.} \qquad
Nenghai Yu\textsuperscript{\rm 1} \\
\textsuperscript{\rm 1}University of Science and Technology of China\quad \
\textsuperscript{\rm 2}Microsoft Cloud AI\quad \ 
\textsuperscript{\rm 3}Simon Fraser University \ \\
{\tt\small\{hqd0037@mail., dlight@mail., zhangwm@, ynh@\}ustc.edu.cn} \\
{\tt\small\{cddlyf@, zhouhang2991@\}gmail.com}
}

\maketitle

\begin{abstract}

Adversary and invisibility are two fundamental but conflict characters of adversarial perturbations. 
Previous adversarial attacks on 3D point cloud recognition have often been criticized for their noticeable point outliers, since they just involve an ``implicit constrain'' like global distance loss in the time-consuming optimization to limit the generated noise. 
While point cloud is a highly structured data format, it is hard to constrain its perturbation with a simple loss or metric properly. 
In this paper, we propose a novel Point-Cloud Sensitivity Map to boost both the efficiency and imperceptibility of point perturbations. 
This map reveals the vulnerability of point cloud recognition models when encountering shape-invariant adversarial noises. 
These noises are designed along the shape surface with an ``explicit constrain'' instead of extra distance loss. 
Specifically, we first apply a reversible coordinate transformation on each point of the point cloud input, to reduce one degree of point freedom and limit its movement on the tangent plane. 
Then we calculate the best attacking direction with the gradients of the transformed point cloud obtained on the white-box model. 
Finally we assign each point with a non-negative score to construct the sensitivity map, which benefits both white-box adversarial invisibility and black-box query-efficiency extended in our work. 
Extensive evaluations prove that our method can achieve the superior performance on various point cloud recognition models, with its satisfying adversarial imperceptibility and strong resistance to different point cloud defense settings. 
Our code is available at: \href{https://github.com/shikiw/SI-Adv}{https://github.com/shikiw/SI-Adv}.

\end{abstract}

\section{Introduction}

\begin{figure}
\centering
\begin{minipage}{0.325\linewidth}
    \centering
    \includegraphics[width=1\linewidth]{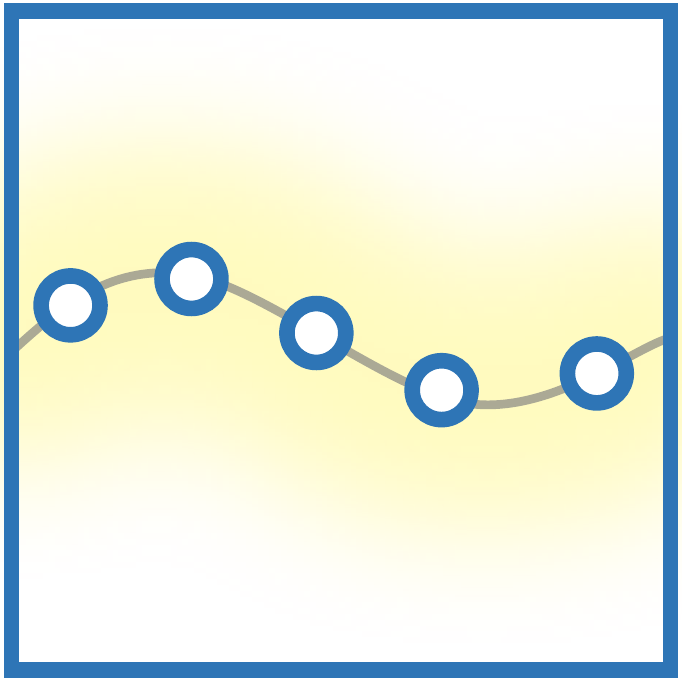}
    \footnotesize(a) Clean PC
\end{minipage}
\hfill
\begin{minipage}{0.325\linewidth}
    \centering
    \includegraphics[width=1\linewidth]{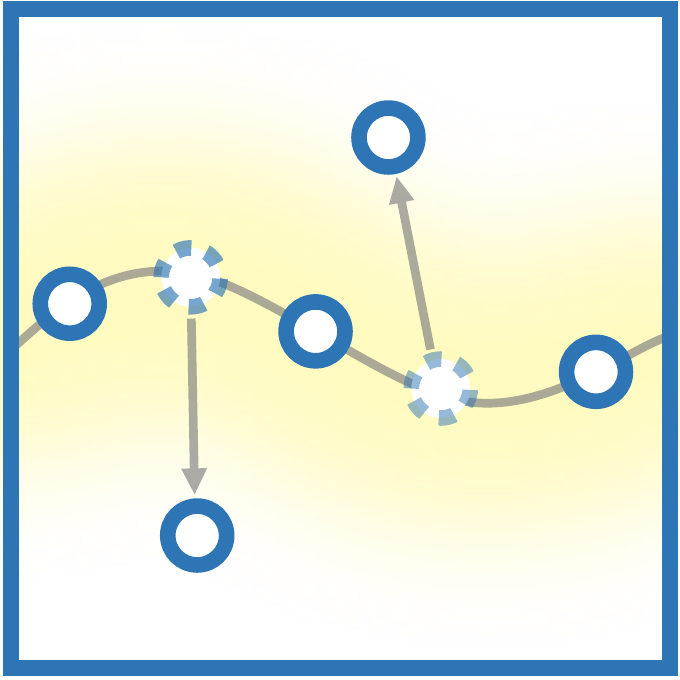}
    \footnotesize(b) Previous Adv 
\end{minipage}
\hfill
\begin{minipage}{0.325\linewidth}
    \centering
    \includegraphics[width=1\linewidth]{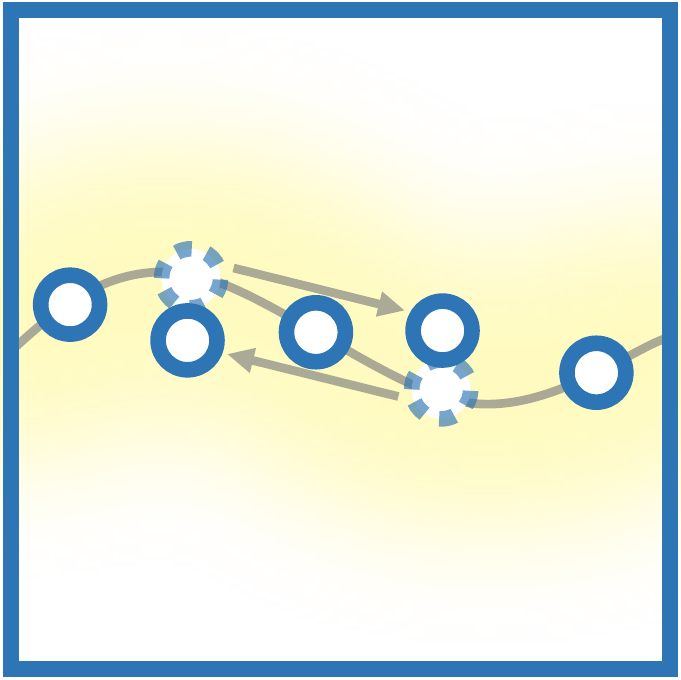}
    \footnotesize(c) Our Adv
\end{minipage}
\caption{The point-cloud shape surface of clean \textbf{(left)}, previous adversarial perturbations \textbf{(middle)} and our perturbation \textbf{(right)}.}
\label{fig:intro_1}
\end{figure}

Deep Neural Networks (DNNs) have contributed a lot to various computer vision tasks with their highly praised performances. However, recent works prove that DNNs are vulnerable to adversarial examples \cite{goodfellow2015explaining,szegedy2013intriguing}, which are generated by adding imperceptible perturbations on clean inputs to mislead the prediction of deep learning models. 
Numerous algorithms have been proposed to implement different kinds of adversarial attacks \cite{goodfellow2015explaining,szegedy2013intriguing,carlini2017towards,gu2015towards,dong18mifgm}. 
In addition to 2D images, the research scope of adversarial attacks has been gradually extended to 3D point clouds. 
As one of the most popular data representations captured by 3D sensors like LiDAR, point clouds are widely used in a variety of 3D computer vision applications, such as autopilot \cite{ZhouT18} and autonomous driving \cite{charles2018frustum}. 
Similar with image recognition models, recent works \cite{SunCCM20,CaoWXYFYCLL21,TuRMLYDCU20,geoa3,Zhou2020lggan,ZhengCYL019,Sun21on} show that point-cloud models are also susceptible to adversarial perturbations.
But unfortunately, existing 3D adversarial attacks either generate the noticeable point perturbation or focus on the white-box case, leading to the poor performance on well-defended models or unseen black-box models.

The motivation for delving into black-box attacks on point cloud recognition is quite clear. 
On the one hand, both industry and academia have attached increasing importance on evaluating the robustness of point cloud models, especially for some security-critical applications like autonomous driving. 
But usually, they can not release the details of the model or algorithms to robustness testers due to the protection agreement of trade secrets. 
Therefore, testers can only conduct black-box attacks to reach the demand of robustness evaluation, which requires the more efficient black-box attack method. 
On the other hand, black-box attacks are obviously more practical but challenging compared with white-box attacks, since the attacker can hardly access to the target model in many realistic scenarios. 
Considering these points, in this paper we adopt a novel shape-invariant perspective to rethink the adversarial imperceptibility and black-box efficiency of adversarial point clouds, based on the following two crucial observations.

\begin{figure}
\centering
\includegraphics[width=1\linewidth]{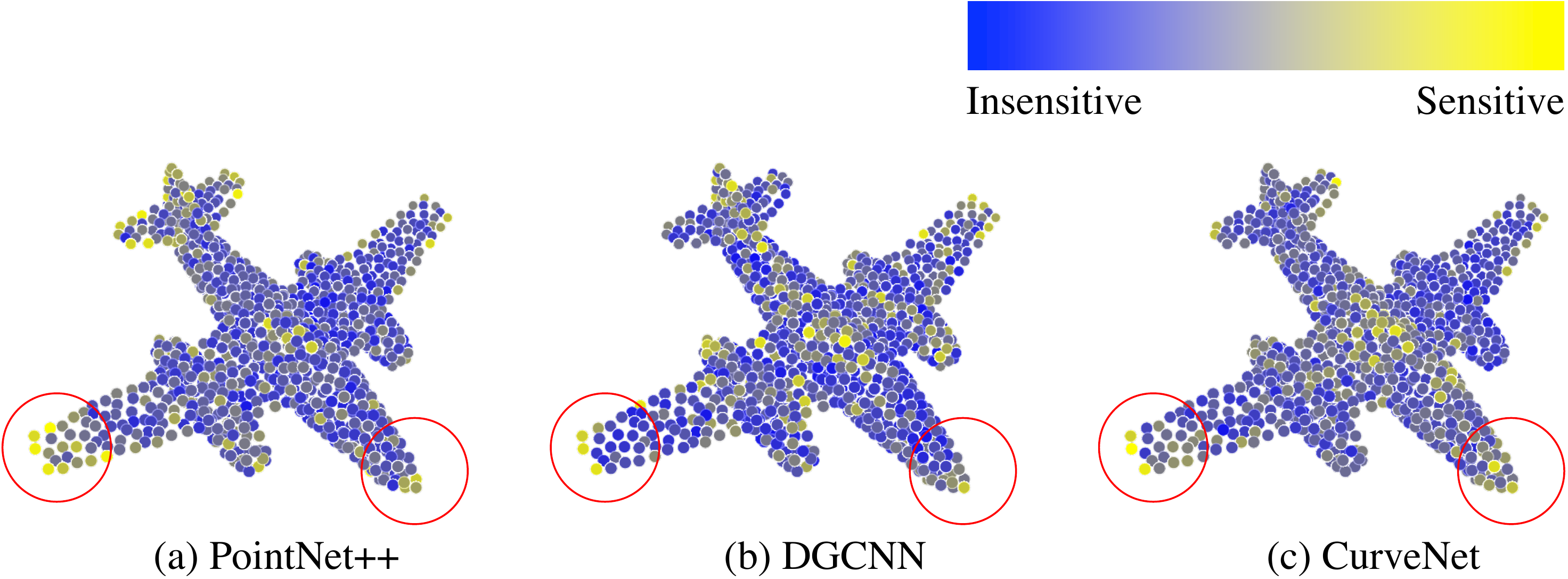}
\caption{Visualization results of the point-cloud sensitivity maps obtained from different 3D recognition models.}
\label{fig:intro_2}
\end{figure}

First, we mathematically analyze the vulnerability of a white-box point cloud recognition model by defining the novel point-cloud sensitivity map. 
This map is constructed by a non-negative score set, where each score measures the variance of the recognition confidence when the corresponding point encounters the imperceptible perturbation. 
To realize such imperceptibility, we intuitively design the point perturbation along the certified optimal direction on the the tangent plane, assuming this slight point shift does not change the shape surface of the point cloud. 
Each point cloud input can obtain its unique sensitivity map on the white-box model, where all points can be ranked in a descending order according to their scores. 
Through iteratively perturbing the top ranked 30\% points, we find that the white-box recognition accuracy can be degraded over 40\% on different models, which demonstrates the existing of shape-invariant white-box attacks.

Second, we discover that these top ranked points obtained from the same point cloud input are similar across different recognition models. 
As visualized in \Fref{fig:intro_2}, the positions of top ranked points almost overlap with each other to a large extent when scored by different 3D classifiers. 
To support this finding more concretely, we look into the 1024-point data samples on the commonly used point-cloud dataset ModelNet40 \cite{Wu2015modelnet} and compare the top ranked points for different recognition models, \ie, PointNet \cite{charles2017pointnet}, PointNet++ \cite{charles2017pointnet++}, DGCNN \cite{dgcnn} and CurveNet \cite{xiang2021curvenet}. 
It shows that the top ranked 30\% points of each point cloud have 75.8\% overlapped between every two models on average and 40.3\% overlapped among every three of them, which are much more than 30\% and 9\% obtained by random sampling respectively. 
Inspired by this internal commonality, we can estimate the top ranked points for the unseen black-box model with a white-box surrogate model to improve the black-box attack ability.

Motivated by the above two observations, our paper further proposes the first query-based black-box attack algorithm on point cloud recognition. 
Through inheriting the ranked sensitivity map prior from white-box surrogate model $\mathcal{H}_w$ to black-box target model $\mathcal{H}_b$, we limit the point movement on the shape surface as shown in \Fref{fig:intro_1}. 
To boost the query efficiency, we combine each point with its certified best direction as a point-wise basis. 
Only one basis is perturbed during each query step according to the sensitivity ranking, in the light of SimBA \cite{Guo2019simba} and LeBA \cite{YangJHNZ20}. 
Experiments on ModelNet40 \cite{Wu2015modelnet} and ShapeNetPart \cite{Yi16shapenet} show that our method achieve the superior performance on various 3D recognition models, which highlights the meaning of the proposed point-cloud sensitivity map.

To summarize, our contributions are three-fold as below.

\begin{itemize}

    \item We propose the point-cloud sensitivity map, for evaluating the variance of recognition confidence when each point encounters the shape-invariant perturbation. 

    \item Guided by the sensitivity map, we propose the strong shape-invariant white-box attack and the first query-based black-box attack on point cloud recognition.
    
    \item Experiments show the efficiency of our attack on various recognition models with the high imperceptibility and the low query cost under black-box settings.
    
\end{itemize}


\section{Related Work}
\subsection{Point Cloud Recognition}

A point cloud is a sparse and unordered point set sampled from object surfaces by 3D sensors. Thanks to the development of deep learning, various methods \cite{charles2017pointnet,charles2017pointnet++,dgcnn,xiang2021curvenet,Xu21paconv,Goyal21simpleview} have been proposed to process 3D point clouds. One of the pioneering works is PointNet \cite{charles2017pointnet}, which directly applies multi-layer perceptron to learn point features and aggregates them with a max-pooling module in an effective way. And its variant, called PointNet++ \cite{charles2017pointnet++}, is a improved hierarchical neural network for better abstracting local features, including single-scale (SSG) and multi-scale (MSG) designs. By exploiting point structures with neighborhood graphs, DGCNN \cite{dgcnn} is another typical work using convolutional networks. In this paper, we mainly evaluate our black-box attack on six representative recognition models.

\subsection{Existing 3D Adversarial Attacks and Defenses}

Current 3D adversarial attack methods can be roughly divided into three categories: point-shifting based \cite{TuRMLYDCU20,Xiang3dadv,Zhou2020lggan,geoa3,HamdiRTG20advpc}, point-adding based \cite{XiangQL19} and point-dropping based \cite{ZhengCYL019}. Following previous works \wrt adversarial attacks on 2D images, a variety 
of point-shifting based methods either utilize the gradient descent or solve an optimization problem like C\&W \cite{carlini2017towards} to perturb point sets and deceive target classifiers. As for point-adding and point-dropping based approaches, they depend on changing the number of input points to realize attacks. For instance, Zheng \etal propose saliency map \cite{Zheng2019saliency} to drop out critical points of the input and get model fooled successfully. Corresponding to attacks, defense methods based on point-cloud preprocessing \cite{zhou2019dupnet}, adversarial training \cite{LiuYS19,Sun21adver}, gather-vector \cite{Dong2020self} or provable defense \cite{liu2021pointguard} have been seriously under-researched.

Since there is still lack of the study on 3D query-based black-box adversarial attack, we delve into this topic based on the proposed sensitivity map in this paper.

\subsection{Query-based Black-box Adversarial Attacks}
Traditional query-based black-box attacks on 2D images can be divided into two categories: decision-based \cite{BrendelRB18,DongSWLL0019} and score-based \cite{ChenZSYH17,TuTC0ZYHC19,Guo2019simba,AndriushchenkoC20,YangJHNZ20}. Both of them mainly rely on the limited times of queries to the unseen black-box model to achieve attacks, without any inner model information or gradient guidance. 
The different things is that, score-based methods utilize the output confidence score, while decision-based methods need the output label from the query feedback. 
In our work, we extend the score-based black-box attack to 3D point-cloud recognition, which is a seriously under-researched field for adversarial attack.

\begin{figure*}[h!]
\begin{minipage}{0.32\linewidth}
    \centering
    \includegraphics[width=1\linewidth]{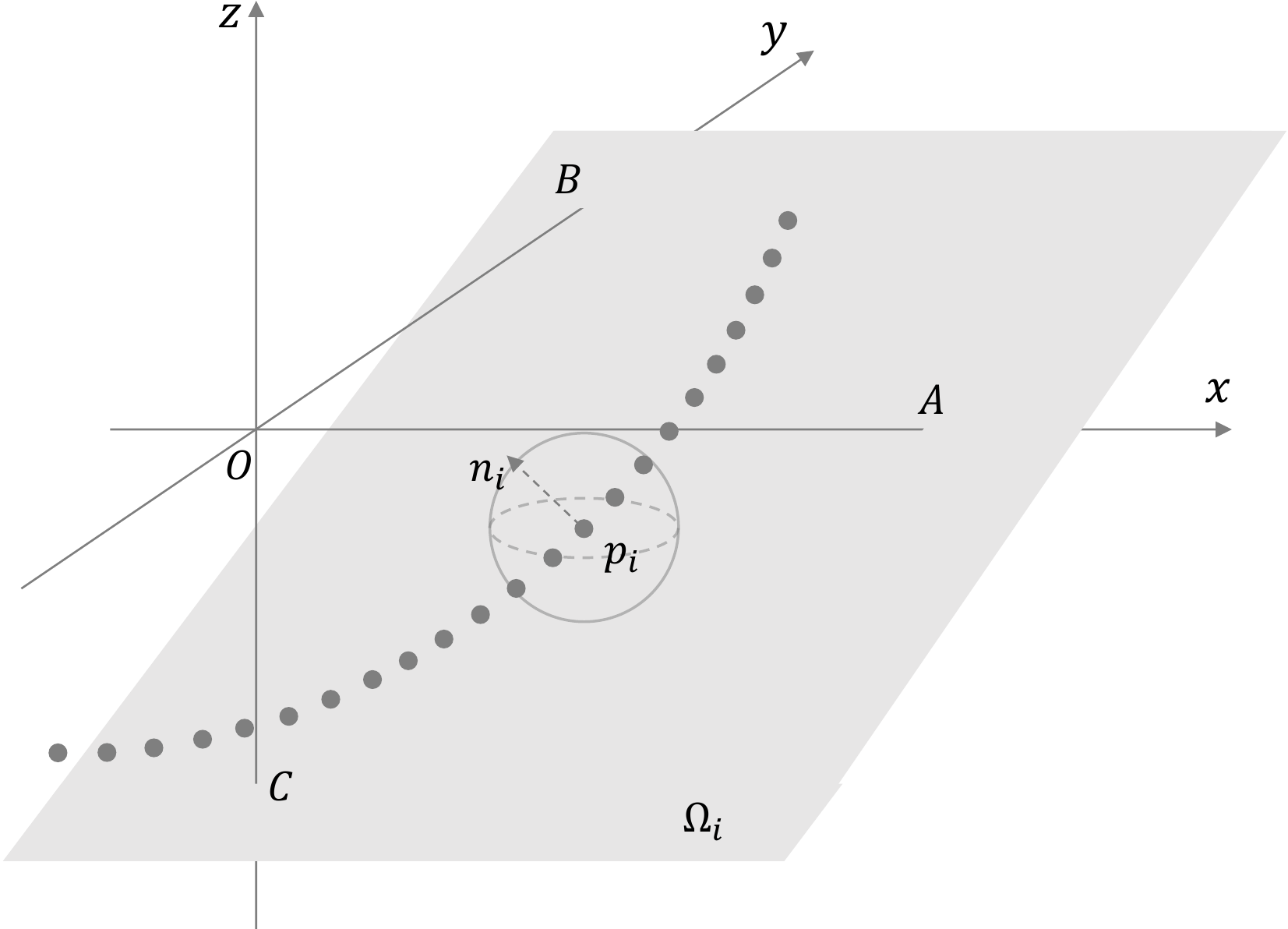}
    \footnotesize(a) Build Tangent Plane
\end{minipage}
\hfill
\begin{minipage}{0.32\linewidth}
    \centering
    \includegraphics[width=1\linewidth]{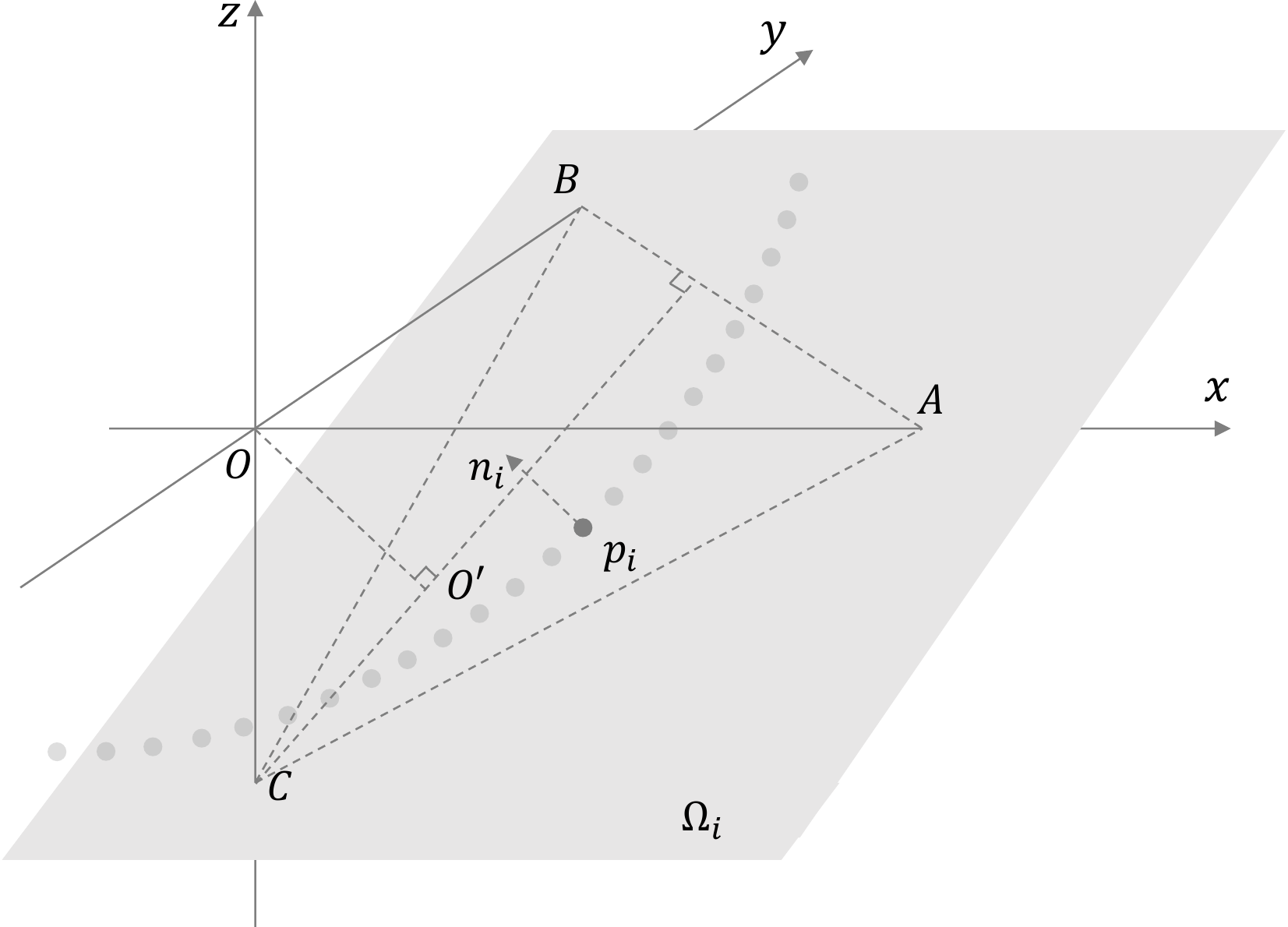}
    \footnotesize(b) Define New Coordinate Basis
\end{minipage}
\hfill
\begin{minipage}{0.32\linewidth}
    \centering
    \includegraphics[width=1\linewidth]{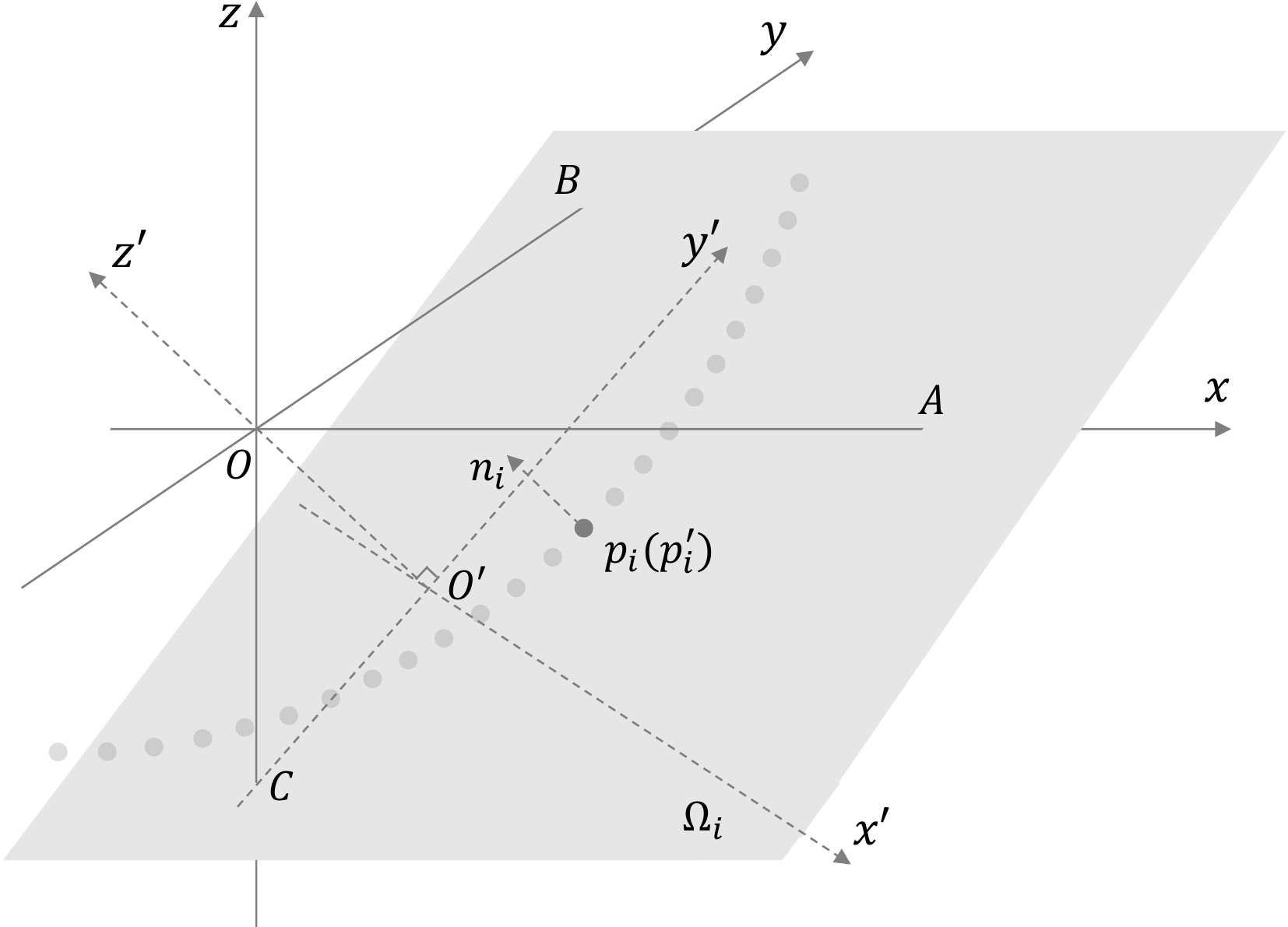}
    \footnotesize(c) Construct New Coordinate System
\end{minipage}
\caption{Schematic diagram of the newly constructed coordinate system for our reversible coordinate transformation.}
\label{fig:pipeline}
\vspace{-0.5em}
\end{figure*}

\section{Preliminaries}

\subsection{Point Cloud Sensitivity Maps}
For a point cloud $\mathcal{P}$ that consists of total $N$ points, we can define it as a set of vectors $\{\boldsymbol{p}_i\}_{i=1}^N$ where $\boldsymbol{p}_i \in \mathbb{R}^3$ denotes the 3D coordinates of each point. 
Also, there exists $k$ nearest neighbors $\mathcal{N}_{\boldsymbol{p}_i}$ for each point $\boldsymbol{p}_i = (p_{i1}, p_{i2}, p_{i3})$, which can help us simulate its normal vector by calculating the $3\times 3$ positive semi-definite matrix \cite{HoppeDDMS92} formulated as
\begin{equation}
    \mathcal{C}_{\boldsymbol{p}_i} = \sum_{\boldsymbol{q} \in \mathcal{N}_{\boldsymbol{p}_i}} (\boldsymbol{q} - \boldsymbol{p}_i) \otimes (\boldsymbol{q} - \boldsymbol{p}_i),
\end{equation}
where $\otimes$ denotes the outer product operation. 
Note that the normal vector $\boldsymbol{n}_i = (n_{i1}, n_{i2}, n_{i3})$ of $\boldsymbol{p}_i$ can be acquired by solving the eigen-decomposition on matrix $\mathcal{C}_{\boldsymbol{p}_i}$ and choosing the last unit eigen-vector.

Once both the shape surface point $\boldsymbol{p}_i$ and its normal vector $\boldsymbol{n}_i$ are determined in the original coordinate system, the corresponding plane that is tangent to the shape surface is also uniquely determined, so called tangent plane $\Omega_i$. 
Since the normal vector $\boldsymbol{n}_i$ is perpendicular to $\Omega_i$, the coordinates of any position $\boldsymbol{m} = (m_1, m_2, m_3)$ on the tangent plane can be easily represented with the following equation: 
\begin{equation}
    (\boldsymbol{m} - \boldsymbol{p}_i) \cdot \boldsymbol{n}_i = \sum_{j=1}^3 (m_j - p_{ij}) n_{ij} = 0
\label{eq:equation2}
\end{equation}
where $\cdot$ denotes the inner product operation. 
As illustrated in \Fref{fig:pipeline}(a), we assume that this tangent plane $\Omega_i$ intersects with the coordinate axes (\ie, $x$ axis, $y$ axis and $z$ axis) of the original coordinate system at position $\boldsymbol{A}$, $\boldsymbol{B}$ and $\boldsymbol{C}$. 
Combined with Eq.(\ref{eq:equation2}), we obtain these three positions as
\begin{equation}
    \boldsymbol{A} (\frac{\boldsymbol{p}_i \cdot \boldsymbol{n}_i}{n_{i1}}, 0, 0), 
    \boldsymbol{B} (0, \frac{\boldsymbol{p}_i \cdot \boldsymbol{n}_i}{n_{i2}}, 0), 
    \boldsymbol{C} (0, 0, \frac{\boldsymbol{p}_i \cdot \boldsymbol{n}_i}{n_{i3}}).
\end{equation}

Obviously, $\boldsymbol{O}$-$\boldsymbol{ABC}$ forms a right angle triangular pyramid since the three axes $x$, $y$ and $z$ are orthogonal with each other, where $\boldsymbol{O}$ denotes the coordinate origin. 
As clarified in \Fref{fig:pipeline}(b), we suppose the projection of origin $\boldsymbol{O}$ on the tangent plane $\Omega_i$ as $\boldsymbol{O'}$. 
Note that $\boldsymbol{OO'}\perp \Omega_i$ and $\boldsymbol{BA} \in \Omega_i$, thus we can get $\boldsymbol{OO'}\perp \boldsymbol{BA}$. 
Likewise, we can also get $\boldsymbol{OC}\perp \boldsymbol{BA}$. 
Based on these two vertical relationship, we can obtain $\boldsymbol{BA}\perp \boldsymbol{COO'}$ where $\boldsymbol{COO'}$ means the plane determined by $\boldsymbol{C}$, $\boldsymbol{O}$ and $\boldsymbol{O'}$. 
Since $\boldsymbol{CO'} \in \boldsymbol{COO'}$, we finally obtain $\boldsymbol{CO'}\perp \boldsymbol{BA}$, which is the fundamental point for us to introduce the following coordinate transformation. 

To limit the movement of $\boldsymbol{p}_i$ on the tangent plane $\Omega_i$ with only two variables, we reduce its one degree of freedom (DoF). 
Intuitively, we consider to leverage a transformed coordinate system to represent this shape surface point, based on the orthogonality among $\boldsymbol{OO'}$, $\boldsymbol{CO'}$ and $\boldsymbol{BA}$. 
Specifically, we assume the direction denoted by $\overrightarrow{\boldsymbol{BA}}$ as the transformed axis $x'$, the direction denoted by $\overrightarrow{\boldsymbol{CO'}}$ as the transformed axis $y'$ and the direction denoted by $\overrightarrow{\boldsymbol{O'O}}$ as the transformed axis $z'$. 
As shown in \Fref{fig:pipeline}(c), the origin of the newly constructed rectangular coordinate system becomes $\boldsymbol{O'}$, with the transformed coordinates of $\boldsymbol{p}_i$ as $\boldsymbol{p}'_i = (p'_{i1}, p'_{i2}, p'_{i3})$. 
In this way, the surface point can be constrained on the tangent plane when we assign the coordinate value on axis $z'$ as zero. 
Therefore, the problem is simplified as formulating the transformation applied for converting the original coordinate system $\boldsymbol{O}$-$xyz$ to the transformed coordinate system $\boldsymbol{O'}$-$x'y'z'$.

\noindent\textbf{Theorem 1.} 
\textit{Assume $f_i: \mathbb{R}^3 \mapsto \mathbb{R}^3$ is the transformation for converting the original coordinate system $\boldsymbol{O}$-$xyz$ to the new coordinate system $\boldsymbol{O'}$-$x'y'z'$. Then $f_i$ is reversibly composited by rotation transformation $f_{ir}$ and translation transformation $f_{it}$, and new $3\times 1$ coordinates $\boldsymbol{p}'_i$ can be converted with original $3\times 1$ coordinates $\boldsymbol{p}_i$ as}
\begin{equation}
    \boldsymbol{p}'_i  = \boldsymbol{R}_i (\boldsymbol{p}_i + \boldsymbol{T}_i ), \quad
    \boldsymbol{p}_i  = \boldsymbol{R}_i^\top \boldsymbol{p}'_i - \boldsymbol{T}_i, 
\label{eq:equation4}
\end{equation}
\textit{where $\boldsymbol{R}_i$ and $\boldsymbol{T}_i$ is the rotation transformation matrix and the translation transformation matrix respectively, \ie, }
\begin{equation}
    \boldsymbol{R}_i = 
    \left(
    \begin{array}{ccc}
    \frac{n_{i2}}{\sqrt{1-n_{i3}^2}} & \frac{-n_{i1}}{\sqrt{1-n_{i3}^2}} & 0 \\
    \frac{n_{i1}n_{i3}}{\sqrt{1-n_{i3}^2}} & \frac{n_{i2}n_{i3}}{\sqrt{1-n_{i3}^2}} & -\sqrt{1-n_{i3}^2} \\
    n_{i1} & n_{i2} & n_{i3}
    \end{array}
    \right), 
\end{equation}
\begin{equation}
    \boldsymbol{T}_i = (\boldsymbol{p}_i \cdot \boldsymbol{n}_i) \boldsymbol{n}_i, 
\end{equation}
\textit{including a boundary case when $|n_{i3}| \rightarrow 1$ that is }
\begin{equation}
    \lim_{|n_{i3}| \rightarrow 1} \boldsymbol{R}_i = 
    \left(
    \begin{array}{ccc}
    1 / \sqrt{2} & -1 / \sqrt{2} & 0 \\
    n_{i3} / \sqrt{2} & n_{i3} / \sqrt{2} & 0 \\
    0 & 0 & n_{i3}
    \end{array}
    \right). 
\end{equation}
\begin{proof}
The proof is given in supplementary materials.
\end{proof}

The above transformation facilitates the convenience of the sensitivity map computation. 
Given a white-box point cloud recognition model $\mathcal{H}_w$ and the point cloud input $\mathcal{P}$, we first apply the reversible transformation elaborated in Eq.(\ref{eq:equation4}) on original coordinates of each point, \ie, $\forall \boldsymbol{p}_i \in \mathcal{P}$, 
\begin{equation}
    \mathcal{P}' = \{\boldsymbol{R}_i (\boldsymbol{p}_i + \boldsymbol{T}_i) \}_{i=1}^N, \quad
    \mathcal{P} = \{\boldsymbol{R}_i^\top \boldsymbol{p}'_i - \boldsymbol{T}_i \}_{i=1}^N,  
\label{eq:equation8}
\end{equation}
then we implement the model inference for $\mathcal{P}$ and compute the max-margin logit loss proposed in C\&W \cite{carlini2017towards}, \ie, 
\begin{equation}
    \mathcal{L} (\mathcal{P}, t; \theta_w) = \mathop{\max} \Big( [\mathcal{H}_w (\mathcal{P})]_t - \mathop{\max}_{j\neq t} [\mathcal{H}_w (\mathcal{P})]_j, 0 \Big),
\end{equation}
where $t$ is the ground-truth label of point cloud $\mathcal{P}$ and $\theta_w$ is the model parameter of white-box model $\mathcal{H}_w$. 
And $[\mathcal{H}_w (\mathcal{P})]_j$ is the probability logit score on class $j$. 
Since the transformation in Eq.(\ref{eq:equation8}) is differentiable, we can easily obtain the gradient map of $\mathcal{P}'$ by back propagation, 
\begin{equation}
    \mathcal{G} = \{\boldsymbol{g}_i\}_{i=1}^N = \nabla_{\mathcal{P}'} \mathcal{L} = \frac{\partial\mathcal{L} (\mathcal{P}, t; \theta_w)}{\partial\mathcal{P}'}, 
\label{eq:equation10}
\end{equation}
where $\boldsymbol{g}_i = (g_{i1}, g_{i2}, g_{i3}) = (\frac{\partial\mathcal{L}}{\partial x'}, \frac{\partial\mathcal{L}}{\partial y'}, \frac{\partial\mathcal{L}}{\partial z'})$ denotes the gradients of loss $\mathcal{L}$ to transformed point coordinates $\boldsymbol{p}'_i$. 

With the guidance of gradient map for the transformed point clouds, the shape-invariant point cloud sensitivity can be defined as: 
\textit{For each point $\boldsymbol{p}_i \in \mathcal{P}$, we define its sensitivity on white-box model $\mathcal{H}_w$ as the maximal variation rate of the logit loss $\mathcal{L}$ when perturbing $\boldsymbol{p}_i$ along any direction on the tangent plane $\Omega_i$.} 
Through assigning the last element $p'_{i3}$ of $\boldsymbol{p'_i}$ as a constant zero and ignoring $g_{i3}$ in the gradients $\boldsymbol{g}_i$, we can limit the point on its tangent plane and measure the aforementioned variation with $\boldsymbol{g}_i = (g_{i1}, g_{i2}, 0)$. 
So the remaining problem is how to find the direction that achieves the maximal variation on the logit loss $\mathcal{L}$.

\noindent\textbf{Theorem 2.}
\textit{Given the function $\mathcal{L}$ and the freedom-reduced ternary variable $\boldsymbol{p}' = (x', y', 0)$ initialized as $(p'_{i1}, p'_{i2}, 0)$, the variation rate of $\mathcal{L}$ is upper bounded at $\sqrt{g_{i1}^2 + g_{i2}^2}$.} 
\begin{proof}
We ignore the constant zero item and shift the variable in the spherical coordinate system where the point is presented as $(x',y',0) = (p'_{i1}+r\cos{\theta}, p'_{i2}+r\sin{\theta}, 0)$, where $\theta$ is the angle between the perturbing direction and the positive half axis of $x'$. 
Hence $r$ can be presented as $r = \sqrt{(x' - p'_{i1})^2 + (y' - p'_{i2})^2}$. 
Based on this, the variation of $\mathcal{L}$ can be regarded as $\nabla_r \mathcal{L}$, upper bounded by
\begin{align}
    \frac{\partial \mathcal{L}}{\partial r} &= 
    \frac{\partial \mathcal{L}}{\partial x'} \frac{\partial x'}{\partial r} + \frac{\partial \mathcal{L}}{\partial y'} \frac{\partial y'}{\partial r} 
    = g_{i1} \cos{\theta} + g_{i2} \sin{\theta} \\ 
    &\leq \sqrt{(g_{i1}^2 + g_{i2}^2)(\cos^2\theta + \sin^2\theta)} 
    = \sqrt{g_{i1}^2 + g_{i2}^2}, 
\end{align}
where we leverage \textit{Cauchy Inequality} and the equal condition of maximum is when angle $\theta = \arctan (g_{i2} / g_{i1})$. 
\end{proof}

Therefore, when we use $\theta$ to represent the perturbing direction on the tangent plane, the sensitivity score of each point can be determined by the maximal variation rate solved above. 
Hereby, the sensitivity map of $\mathcal{P}$ is defined as
\begin{equation}
    \mathcal{S} = \{\boldsymbol{s}_i\}_{i=1}^N, \quad \text{subject to: } \boldsymbol{s}_i = \sqrt{g_{i1}^2 + g_{i2}^2}
\label{eq:equation13}
\end{equation}

\subsection{Shape-invariant White-box Attack}

With the same definition on point cloud $\mathcal{P}\in \mathbb{R}^{N\times 3}$, white-box model $\mathcal{H}_w$ and gradient map $\mathcal{G} = \{(g_{i1}, g_{i2}, 0)\}_{i=1}^N$, the shape-invariant coordinate transformation denoted in Eq.(\ref{eq:equation8}) are applied in our white-box attack. 
Specifically, we first perform gradient-based optimization on the transformed point cloud $\mathcal{P}'$ with white-box model $\mathcal{H}_w$ to generate the transformed adversarial point cloud $\tilde{\mathcal{P}}'$. 
Then we apply the inverse transform mentioned in Eq.(\ref{eq:equation8}) to get adversarial point cloud $\tilde{\mathcal{P}}$. 
Inspired by I-FGSM \cite{gu2015towards}, we implement the adversarial example generation in an iterative style, where rotation matrix gather $\mathcal{R} = \{\boldsymbol{R}_i\}_{i=1}^N$ and translation matrix gather $\mathcal{T} = \{\boldsymbol{T}_i\}_{i=1}^N$ are iteratively updated in each step $t$ to inherit the shape invariance. 
To simplify the formulation, we first define the following mappings: 
\begin{equation}
    F_t (\mathcal{P}) = \mathcal{R}_t (\mathcal{P} + \mathcal{T}_t),\quad F'_t (\mathcal{P}') = \mathcal{R}_t^\top \mathcal{P}' - \mathcal{T}_t,
\end{equation}
where $\mathcal{R}^\top = \{ \boldsymbol{R}_i^\top\}_{i=1}^N$. 
Then in each step $t$, our untargeted adversarial example generation can be formulated as
\begin{align}
    \tilde{\mathcal{P}}'_{t+1} &= \tilde{\mathcal{P}}'_t - \beta \cdot \frac{\nabla_{\tilde{\mathcal{P}}'} \mathcal{L} (F'_t(\tilde{\mathcal{P}}'_t), t; \theta_w)}{\| \nabla_{\tilde{\mathcal{P}}'} \mathcal{L} (F'_t(\tilde{\mathcal{P}}'_t), t; \theta_w) \|}, \\
    \tilde{\mathcal{P}}_{t+1} &= Clip (F'_t (\tilde{\mathcal{P}}'_{t+1})), \\
    \tilde{\mathcal{P}}'_{t+1} &= F_{t+1} (\tilde{\mathcal{P}}_{t+1}), 
\end{align}
where $\beta$ means the step size and $Clip$ means the $l_\infty$-norm constraint. 
When the generation is completed, the adversarial point cloud $\tilde{\mathcal{P}}$ can fool $\mathcal{H}_w$ with satisfying visual quality.

\begin{algorithm}[h!]
    \caption{Shape-invariant Query-based Attack}
    \label{alg:Algo1}
    \LinesNumbered
    \KwIn{point-cloud input $(\mathcal{P},l)$, black-box model $\mathcal{H}_b$, surrogate model $\mathcal{H}_w$ and step size $\epsilon$.}
    \KwOut{adversarial point cloud $\tilde{\mathcal{P}}$}
    Initialize the perturbation $\delta = 0$ \\
    Initialize the prediction pool $\mathbf{p}_c = [\mathcal{H}_b (\mathcal{P})]_c$ \\
    Transform $\mathcal{P}$ to $\mathcal{P}'$ \hfill $\lhd$ \text{Eq.(\ref{eq:equation8})} \\
    Compute gradient map $\mathcal{G}$ of $\mathcal{P}'$ on $\mathcal{H}_w$ \hfill $\lhd$ \text{Eq.(\ref{eq:equation10})} \\
    Compute and rank sensitivity map $\mathcal{S}$ \hfill $\lhd$ \text{Eq.(\ref{eq:equation13})} \\
    \While{$l = \mathop{\arg\max}_c \mathbf{p}_c$ \textbf{and} $\mathcal{S} \neq \varnothing$}{
        Pick top ranked $q \in \mathcal{S}$ and $\mathcal{S} = \mathcal{S}\backslash \{q\}$ \\
        Get its direction $\theta = \arctan (g_{i2} / g_{i1})$ \\
        Compute basis $q = q\cdot(\cos\theta, \sin\theta, 0)$ \\
        \For{$\alpha \in \{\epsilon, -\epsilon\}$}{
            Reverse $\mathcal{P}'+\delta+\alpha q$ to $\mathcal{P}_{inp}$ \hfill $\lhd$ \text{Eq.(\ref{eq:equation8})} \\
            Get prediction $\mathbf{p}_c' = [\mathcal{H}_b (\mathcal{P}_{inp})]_c$ \\
            \If{$\mathbf{p}_l' < \mathbf{p}_l$}{
                Update $\delta = \delta + \alpha q$ \\
                New prediction pool $\mathbf{p}_c = \mathbf{p}_c'$
            }
        }
    }
    \Return $\tilde{\mathcal{P}} = \mathcal{P}_{inp}$
\end{algorithm}

\subsection{Shape-invariant Black-box Attack}
Given a clean point cloud $\mathcal{P}$ and its ground-truth label $l$, the goal of query-based black-box attack is to compute the constrained adversarial perturbation $\delta$ through fully utilizing the limited times of queries to black-box model $\mathcal{H}_b$. 
The queries provide the output logits, \ie, probability scores predicted on each class. 
In our method, we assume the attacker can adopt a white-box surrogate model $\mathcal{H}_w$ to compute the sensitivity map $\mathcal{S}$ of $\mathcal{P}$. 
The non-negative scores on $\mathcal{S}$ are ranked in a descending order, so that the most sensitive points could be ranked first. 
For a black-box model $\mathcal{H}_b$, we give priority to attacking these top ranked points through regarding each point as a basic basis, as done by SimBA \cite{Guo2019simba}. 
Specifically, in each query step, we only perturb one point along or back along the corresponding direction $\theta$ obtained in sensitivity map calculation, since we do not know whether the gradient value in black-box model $\mathcal{H}_b$ is postive or negative. 
The algorithm is detailed in Algo \ref{alg:Algo1}.

\noindent\textbf{Theorem 3.}
\textit{For our query-based black-box adversarial attack with maximal steps $T$ and fixed step size $\epsilon$, the $l_2$ norm of perturbation on $\mathcal{P}$ is upper bounded at $\sqrt{T}\epsilon$.}
\begin{proof}
Since the perturbation searching is to add or subtract on $\mathcal{P}'+\delta$, each basis $q$ will be discarded if both $+\epsilon q$ and $-\epsilon q$ fail to reduce the confidence $\mathbf{P}_l$. 
Thus we consider $\alpha_t \in \{\epsilon, 0, -\epsilon\}$ for each step $t$ and describe the perturbation on $\mathcal{P}'$ with a sum as $\sum_{t=1}^T \alpha_t q_t$, where $q_t \in \mathcal{S}$ is the searched basis during each step and the total step number $T$ is just the amount of $q_t$. Note that these bases are orthogonal with each other, \ie, $q_i^{\top} q_j = 0$, $\forall q_i, q_j \in \mathcal{S}, i \neq j$. 
So combined with $\|\mathcal{R}^\top\|_2^2 = \{1\}_{i=1}^N$, the $l_2$-norm of the perturbation on $\mathcal{P}$ can be upper bounded with this orthogonal property by
\begin{align}
    &\|\delta_\mathcal{P}\|_2^2 = \|(\mathcal{R}^\top \tilde{\mathcal{P}}' - \mathcal{T}) - (\mathcal{R}^\top \mathcal{P}' - \mathcal{T}))\|_2^2 \\
    &= \|\mathcal{R}^\top (\tilde{\mathcal{P}}' - \mathcal{P}') \|_2^2 \leq \|\mathcal{R}^\top\|_2^2 \Big\|\sum_{t=1}^T \alpha_t q_t\Big\|_2^2 \\
    &= \Big\lVert\sum_{t=1}^T \alpha_t q_t\Big\rVert_2^2 = \sum_{t=1}^T\|\alpha_t q_t\|_2^2 + \sum_{i\neq j} \|\alpha_i q_i \alpha_j q_j\|_2^2 \\
    &= \sum_{t=1}^T\|\alpha_t q_t\|_2^2 = \sum_{t=1}^T\alpha_t^2\|q_t\|_2^2 \leq \sum_{t=1}^T \epsilon^2 \cdot 1 = T\epsilon^2.
\end{align}

Thus the tight upper bound $\tau$ of adversarial perturbation $\delta_\mathcal{P}$ is relevant to the maximum steps $T$, \ie, $\tau=\sqrt{T}\epsilon$.
\end{proof}


\begin{table*}[t]
	\footnotesize
    \centering
    \setlength{\tabcolsep}{1mm}{
    \begin{tabular}{lcp{10mm}<{\centering}p{10mm}<{\centering}p{10mm}<{\centering}p{10mm}<{\centering}p{10mm}<{\centering}p{10mm}<{\centering}p{10mm}<{\centering}p{10mm}<{\centering}p{10mm}<{\centering}p{10mm}<{\centering}p{10mm}<{\centering}p{10mm}<{\centering}}
        \toprule
        \multirow{3}{*}{Attack}
        & \multirow{3}{*}{Defense}
        & \multicolumn{4}{c}{PointNet \cite{charles2017pointnet}} 
        & \multicolumn{4}{c}{DGCNN \cite{dgcnn}}
        & \multicolumn{4}{c}{CurveNet \cite{xiang2021curvenet}}
        \\
        \cmidrule(lr){3-6}\cmidrule(lr){7-10}\cmidrule(lr){11-14}
        & & ASR$\uparrow$ & CD$\downarrow$ & HD$\downarrow$ & A.T$\downarrow$ & ASR$\uparrow$ & CD$\downarrow$ & HD$\downarrow$ & A.T$\downarrow$ & ASR$\uparrow$ & CD$\downarrow$ & HD$\downarrow$ & A.T$\downarrow$ 
        \\
        & & (\%) & ($10^{-4}$) & ($10^{-2}$) & (s) & (\%) & ($10^{-4}$) & ($10^{-2}$) & (s) & (\%) & ($10^{-4}$) & ($10^{-2}$) & (s) 
        \\
        \midrule
        I-FGM \cite{gu2015towards} & - & 100.0 & 6.96 & 3.04 & 1.17 & 99.5 & 16.85 & 2.39 & 2.02 & 100.0 & 15.61 & 2.89 & \textbf{10.77} 
        \\
        MI-FGM \cite{dong18mifgm} & - & 99.5 & 35.99 & 4.16 & 1.31 & 95.9 & 120.10 & 4.99 & 2.14 & 99.0 & 119.17 & 5.10 & 10.88  
        \\
        PGD \cite{Madry18adversarial} & - & \textbf{100.0} & 7.00 & 3.05 & \textbf{1.17} & 99.4 & 16.77 & 2.40 & \textbf{2.02} & 100.0 & 15.32 & 2.86 & 10.78  
        \\
        \midrule
        3d-Adv \cite{Xiang3dadv} & - & 99.9 & 3.25 & 2.11 & 4.94 & \textbf{100.0} & 10.12 & 2.46 & 18.73 & 100.0 & 7.48 & 3.51 & 116.24  
        \\
        AdvPC \cite{HamdiRTG20advpc} & - & 99.8 & 16.57 & 3.43 & 2.90 & 98.8 & 14.35 & 1.48 & 8.10 & 99.9 & 18.36 & 2.78 & 64.76 
        \\
        \textbf{Ours} & - & 99.8 & \textbf{2.15} & \textbf{2.04} & 1.32 & 99.9 & \textbf{6.33} & \textbf{1.27} & 3.87 & \textbf{100.0} & \textbf{6.25} & \textbf{1.99} & 21.53  
        \\
        \bottomrule
    \end{tabular}
    }
    \caption{Quantitative comparison between our white-box shape-invariant attack and existing white-box attacks in terms of attack success rate (ASR), Chamfer distance (CD), Hausdorff distance (HD) and average time budget for each adversarial point cloud generation (A.T), where CD is multiplied by $10^4$ and HD is multiplied by $10^2$ for better comparison.}
    \label{tab:table1}
\end{table*}

\section{Experiments}

\subsection{Setup}

\noindent\textbf{Dataset and Data Augmentation.} 
We use ModelNet40 \cite{Wu2015modelnet} and ShapeNetPart \cite{Yi16shapenet} to evaluate the attack performance. 
ModelNet40 consists of 12,311 CAD models from 40 object categories, in which 9,843 models are intended for training and the other 2,468 for testing. 
ShapeNetPart consists of 16,881 pre-aligned shapes from 16 categories, split into 12,137 for training and 2,874 for testing. 
In order to standardize and unify the verification for our method, we uniformly sample 1,024 points from the surface of each object and rescale them into a unit cube, as done by Qi \etal \cite{charles2017pointnet}. 
Besides, several kinds of data augmentation such as random point-dropping and rotation are applied to preprocess all of point clouds in the training set.

\noindent\textbf{Applied Models.} 
Our approach is evaluated on six categories of the most popular 3D recognition models, \ie, PoinetNet \cite{charles2017pointnet}, PointNet++ (MSG) \cite{charles2017pointnet++}, DGCNN \cite{dgcnn}, PAConv \cite{Xu21paconv}, SimpleView \cite{Goyal21simpleview} and CurveNet \cite{xiang2021curvenet}. 
In the manuscript, we mainly report the experimental results on ModelNet40 \cite{Wu2015modelnet}, in which we choose DGCNN \cite{dgcnn} as the surrogate model for black-box attack evaluation. 
The results on ShapeNetPart \cite{Yi16shapenet} or with PointNet \cite{charles2017pointnet} as the surrogate model can be found in supplementary materials.

\begin{figure}
\vspace{-1em}
\centering
\includegraphics[width=1\linewidth]{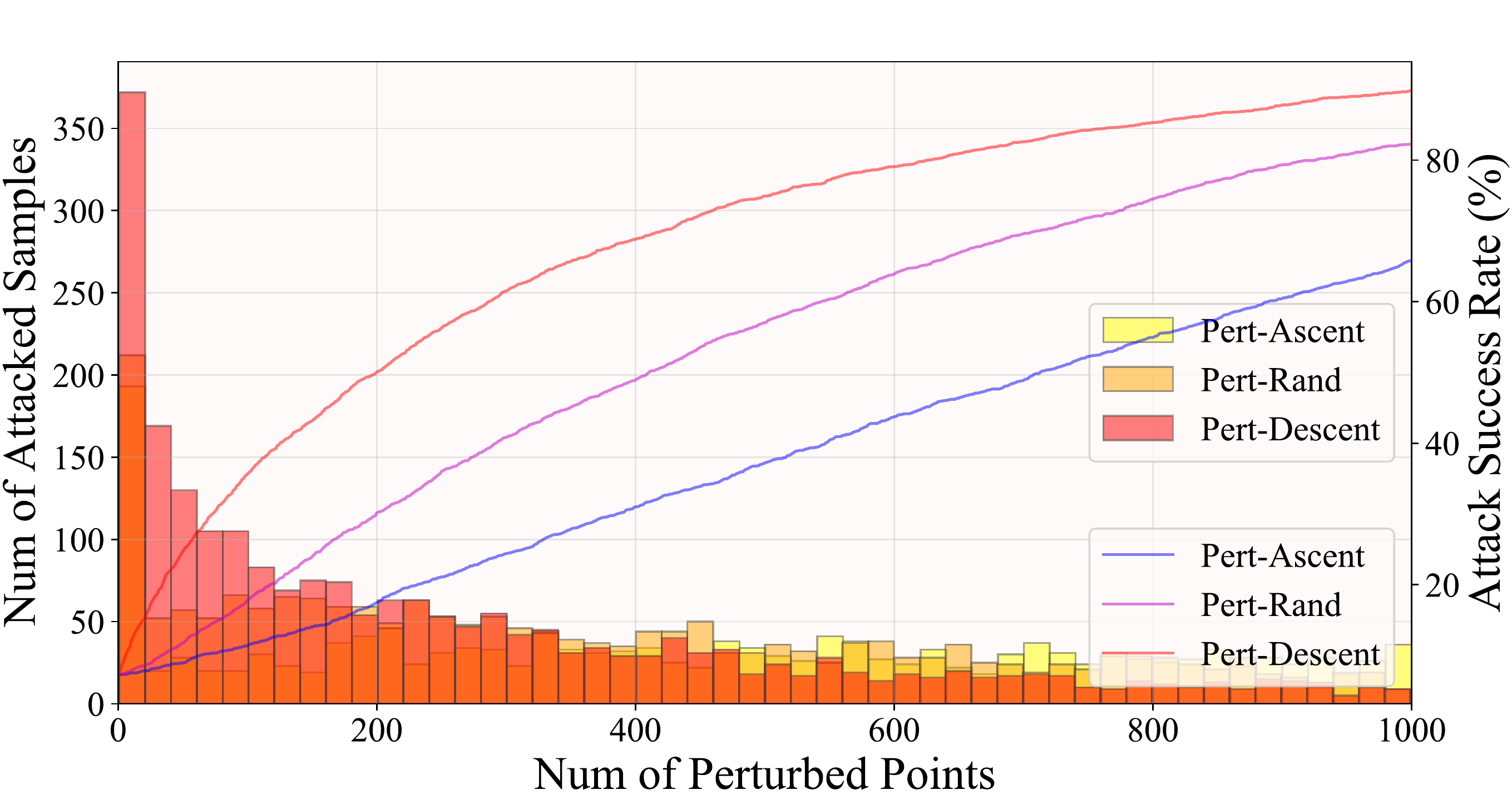}
\caption{Attack success rate (\textbf{curve}) and successfully attacked sample numbers (\textbf{histogram}) when perturbing different number of input points on PAConv, the step size is unified as 0.03.}
\label{fig:query_asr}
\end{figure}

\noindent\textbf{Evaluation Metrics.} 
To verify the imperceptibility of the generated adversarial point clouds, we compute $l_2$-norm distance, Chamfer distance (CD) \cite{FanSG17} and Hausdorff distance (HD) \cite{Taha2015hausdorff} between the original point cloud and the adversarial output to measure the modification caused by attack methods. 
To investigate the attack performance quantitatively, attack success rate (ASR) is further defined as the ratio of perturbed examples that are incorrectly classified by the black-box model. 
Besides, both query numbers and time cost are used for measuring the algorithm efficiency.

\subsection{Efficiency of Sensitivity Maps}

We first conduct the experiment to verify the theoretical validity of the proposed point-cloud sensitivity map. 
In practice, we follow different orders to perturb each single point of the point cloud input on PAConv \cite{Xu21paconv}, 
where the perturbation direction is along the shape surface and the step size is unified as 0.03.
Here ``Pert-Descent'' means following the descending order of sensitivity maps, and ``Pert-Ascent'' means following the ascending order by contrast. 
We also validate the attack performance on the random order denoted by ``Pert-Rand''. 
As indicated in \Fref{fig:query_asr}, the recognition accuracy gets much more degradation in ``Pert-Descent'', proving the effectiveness brought by the proposed sensitivity map and its descending score ranking.

\begin{table}[t]
	\footnotesize
    \centering
    \setlength{\tabcolsep}{1mm}{
    \begin{tabular}{lp{14mm}<{\centering}p{14mm}<{\centering}p{14mm}<{\centering}p{14mm}<{\centering}}
    \toprule
    Defense
    & 3d-Adv \cite{Xiang3dadv}
    & AdvPC \cite{HamdiRTG20advpc}
    & GeoA$^3$ \cite{geoa3}
    & \textbf{Ours}
    \\
    \midrule
    - & 99.9 & 99.8 & \textbf{100.0} & 99.8
    \\
    SOR \cite{zhou2019dupnet} & 12.0 & 33.6 & 69.6 & \textbf{97.4}
    \\
    DUP-Net \cite{zhou2019dupnet} & 12.3 & 29.0 & 73.7 & \textbf{95.8}
    \\
    AT \cite{Madry18adversarial} & 94.8 & 78.6 & \textbf{99.6} & 90.0
    \\
    \bottomrule
    \end{tabular}
    }
    \caption{Attack success rate (ASR) results of different white-box point cloud attacks on PointNet equipped with various defenses.}
    \label{tab:table2}
\end{table}

\begin{figure*}[h!]
\begin{minipage}{0.245\linewidth}
    \centering
    \includegraphics[width=1\linewidth]{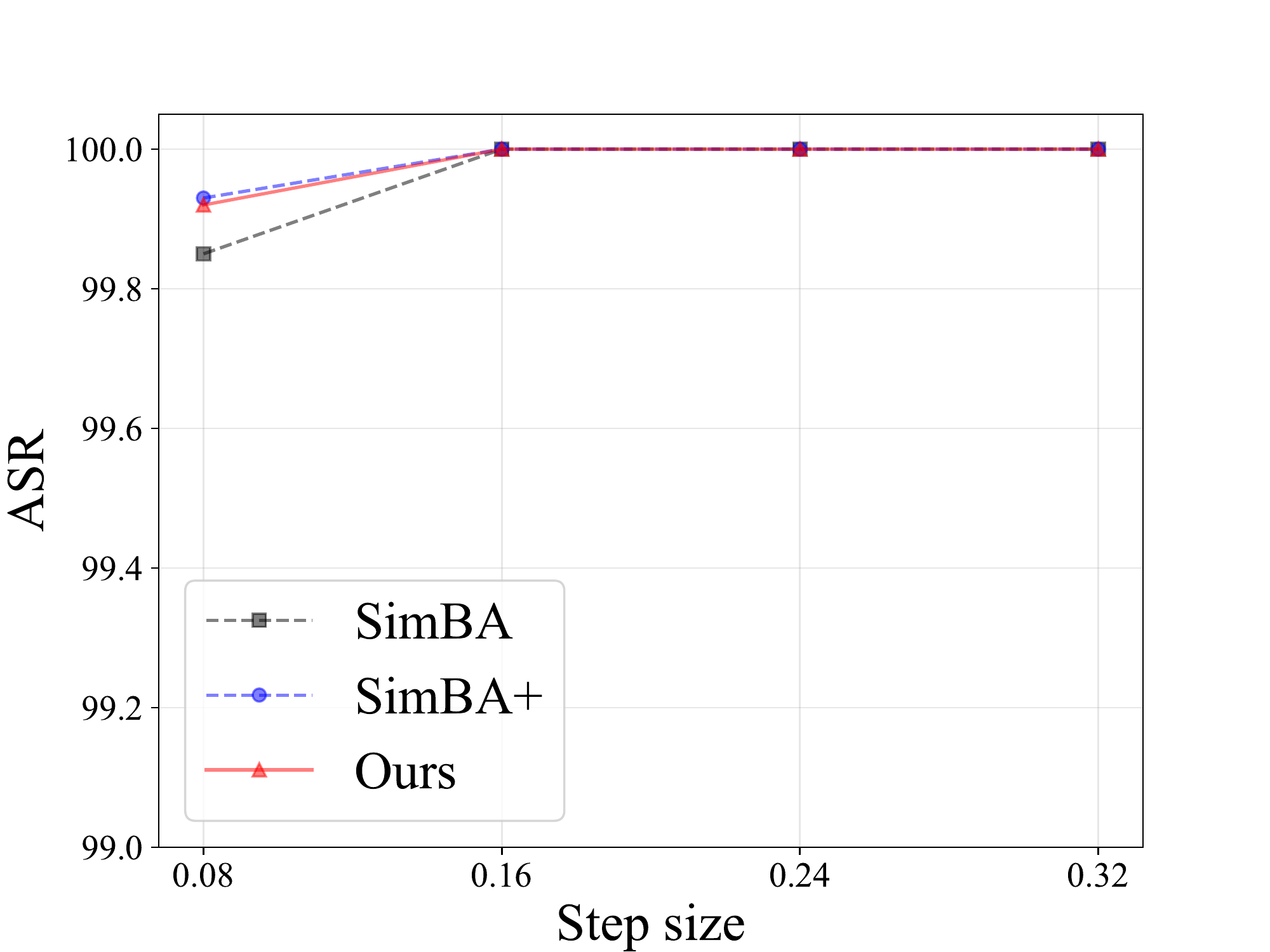}
\end{minipage}
\hfill
\begin{minipage}{0.245\linewidth}
    \centering
    \includegraphics[width=1\linewidth]{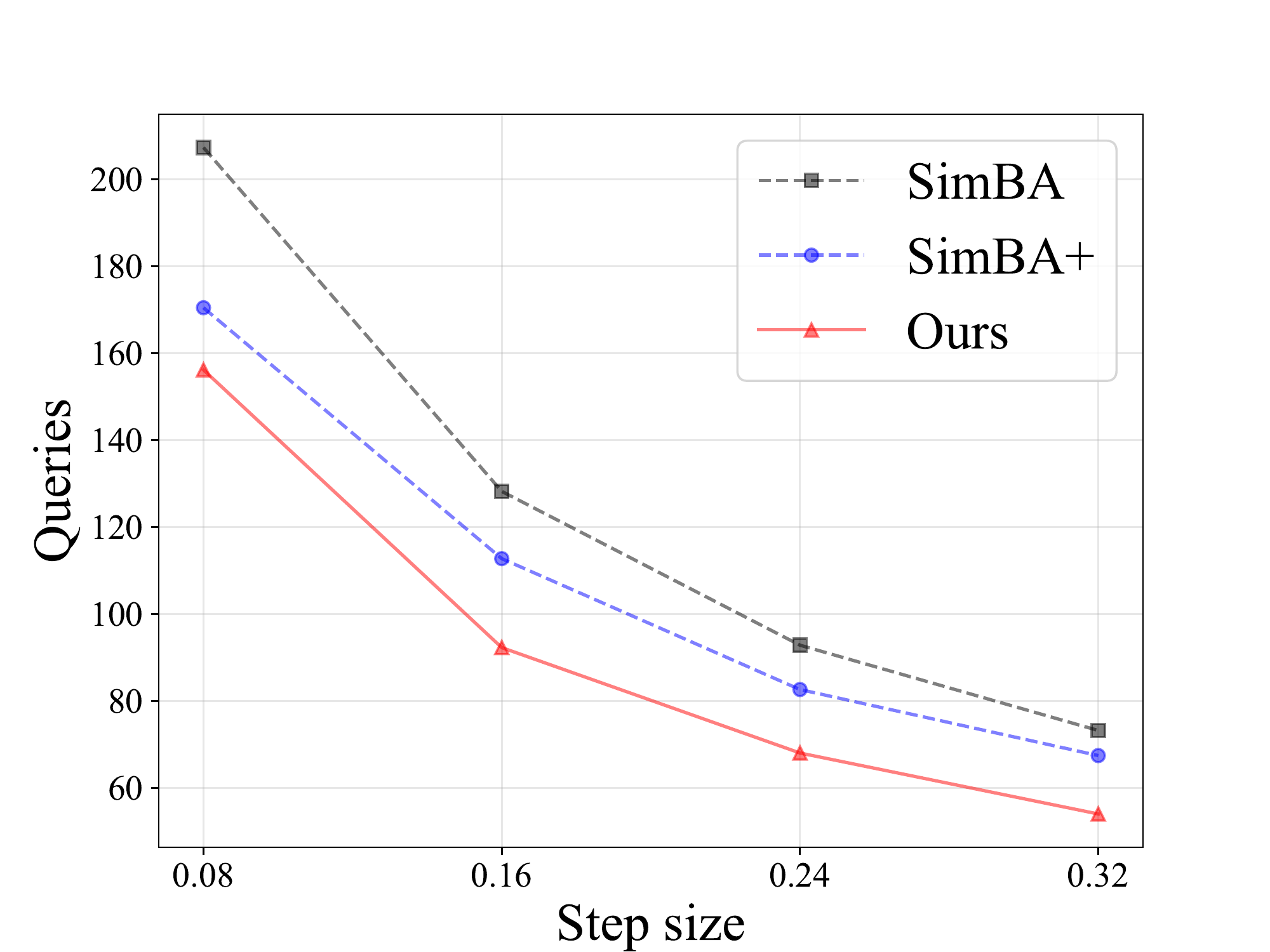}
\end{minipage}
\hfill
\begin{minipage}{0.245\linewidth}
    \centering
    \includegraphics[width=1\linewidth]{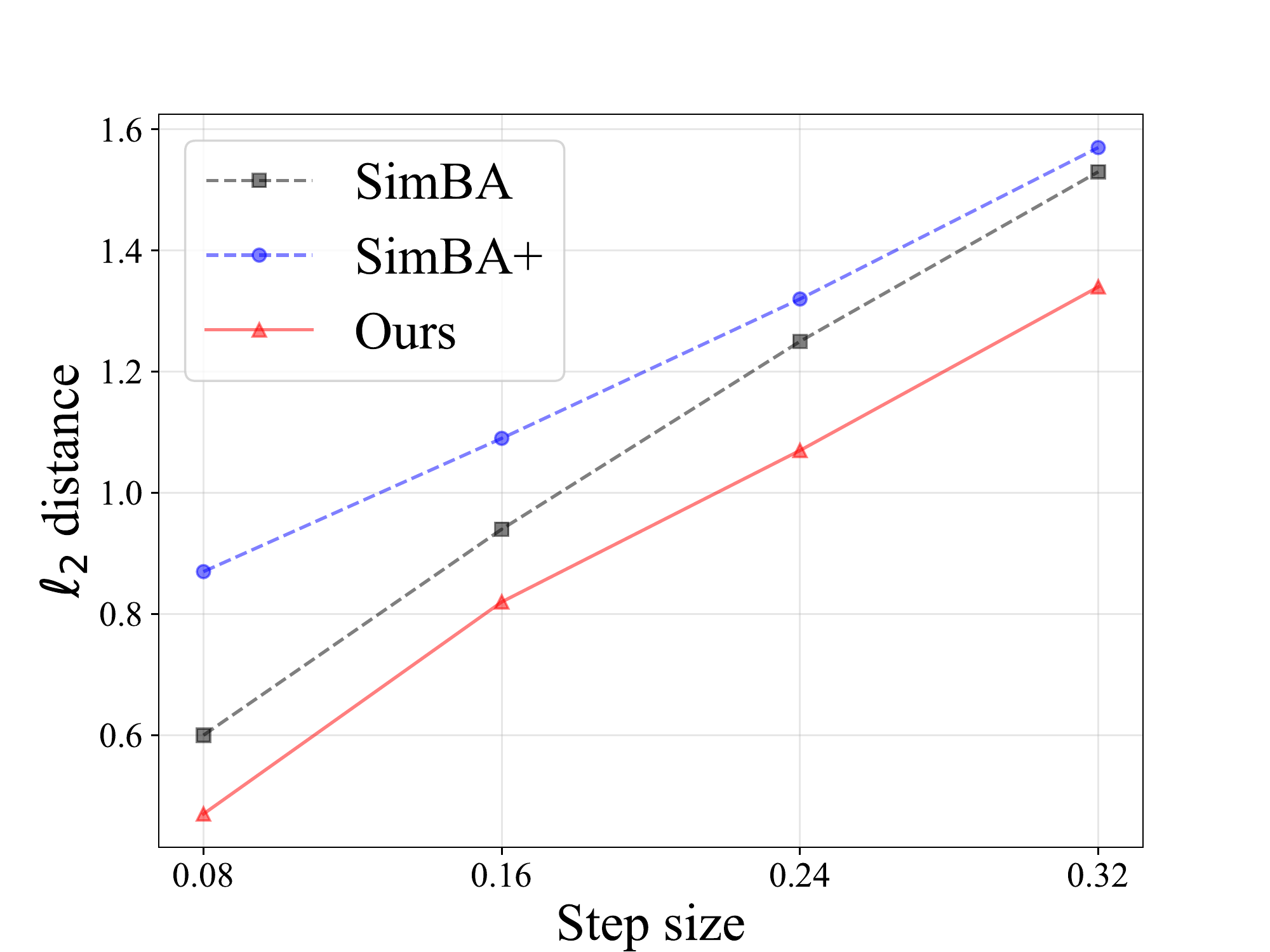}
\end{minipage}
\hfill
\begin{minipage}{0.245\linewidth}
    \centering
    \includegraphics[width=1\linewidth]{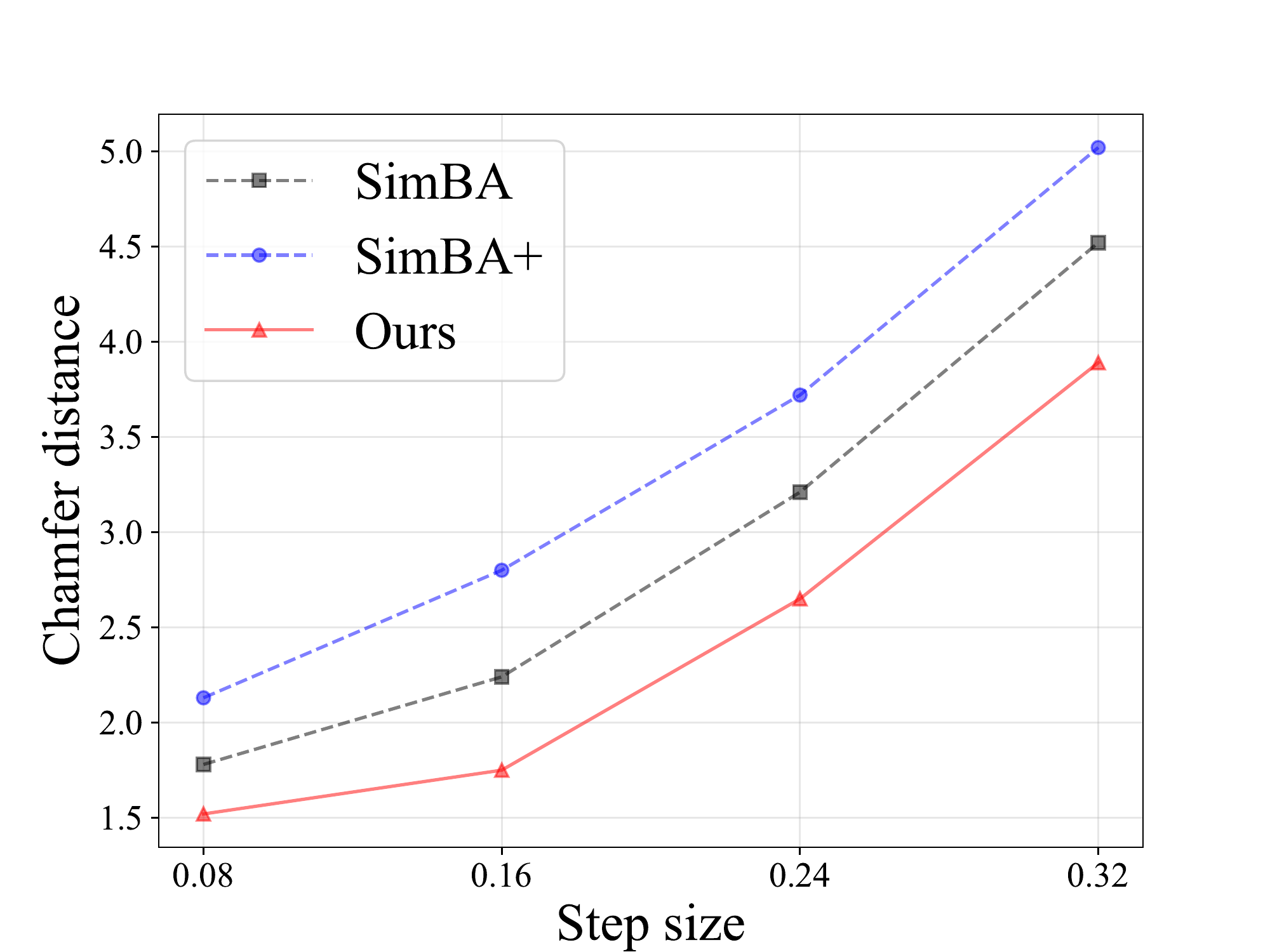}
\end{minipage}
\caption{Black-box performance when applying different attack step size, in terms of attack success rate (ASR), query cost, $l_2$-norm distance and Chamfer distance results between adversarial point clouds and originals.}
\label{fig:graph_paconv}
\end{figure*}

\begin{table*}[h!]
	\footnotesize
    \centering
    \setlength{\tabcolsep}{1mm}{
    \begin{tabular}{lp{8.5mm}<{\centering}p{8.5mm}<{\centering}p{8.5mm}<{\centering}p{8.5mm}<{\centering}p{8.5mm}<{\centering}p{8.5mm}<{\centering}p{8.5mm}<{\centering}p{8.5mm}<{\centering}p{8.5mm}<{\centering}p{8.5mm}<{\centering}p{8.5mm}<{\centering}p{8.5mm}<{\centering}p{8.5mm}<{\centering}p{8.5mm}<{\centering}p{8.5mm}<{\centering}}
        \toprule
        \multirow{3}{*}{Attack}
        & \multicolumn{5}{c}{PointNet++ \cite{charles2017pointnet++}} 
        & \multicolumn{5}{c}{PAConv \cite{Xu21paconv}}
        & \multicolumn{5}{c}{CurveNet \cite{xiang2021curvenet}}
        \\
        \cmidrule(lr){2-6}\cmidrule(lr){7-11}\cmidrule(lr){12-16}
        & ASR$\uparrow$ & A.Q$\downarrow$ & CD$\downarrow$ & HD$\downarrow$ & A.T$\downarrow$ & ASR$\uparrow$ & A.Q$\downarrow$ & CD$\downarrow$ & HD$\downarrow$ & A.T$\downarrow$ & ASR$\uparrow$ & A.Q$\downarrow$ & CD$\downarrow$ & HD$\downarrow$ & A.T$\downarrow$ 
        \\
        & (\%) & (times) & ($10^{-4}$) & ($10^{-2}$) & (s) & (\%) & (times) & ($10^{-4}$) & ($10^{-2}$) & (s) & (\%) & (times) & ($10^{-4}$) & ($10^{-2}$) & (s) 
        \\
        \midrule
        SimBA \cite{Guo2019simba} & 94.9 & 683.8 & 3.05 & 4.80 & 112.04 & 100.0 & 73.2 & 4.51 & 4.93 & 0.36 & 100.0 & 114.5 & 4.67 & 4.93 & 13.51
        \\
        SimBA+ \cite{YangJHNZ20} & 95.5 & 603.4 & 3.41 & 8.21 & 99.57 & 100.0 & 67.4 & 5.02 & 9.58 & 0.36 & \textbf{100.0} & 98.9 & 5.29 & 9.76 & 12.94 
        \\
        \textbf{Ours} & \textbf{95.8} & \textbf{417.0} & \textbf{2.66} & \textbf{4.44} & \textbf{76.81} & \textbf{100.0} & \textbf{53.9} & \textbf{3.89} & \textbf{4.58} & \textbf{0.31} & 99.9 & \textbf{81.5} & \textbf{3.97} & \textbf{4.59} & \textbf{8.77} 
        \\
        \bottomrule
    \end{tabular}
    }
    \caption{Quantitative comparison between our black-box shape-invariant attack and different black-box attacks with unified step size 0.32 on attack success rate (ASR), average query cost (A.Q), Chamfer distance (CD), Hausdorff distance (HD) and average time budget (A.T).}
    \label{tab:table3}
\end{table*}

\begin{table*}[h!]
	\footnotesize
    \centering
    \setlength{\tabcolsep}{1mm}{
    \begin{tabular}{lp{8.5mm}<{\centering}p{8.5mm}<{\centering}p{8.5mm}<{\centering}p{8.5mm}<{\centering}p{8.5mm}<{\centering}p{8.5mm}<{\centering}p{8.5mm}<{\centering}p{8.5mm}<{\centering}p{8.5mm}<{\centering}p{8.5mm}<{\centering}p{8.5mm}<{\centering}p{8.5mm}<{\centering}p{8.5mm}<{\centering}p{8.5mm}<{\centering}p{8.5mm}<{\centering}}
        \toprule
        \multirow{3}{*}{Defense}
        & \multicolumn{5}{c}{PAConv \cite{Xu21paconv}} 
        & \multicolumn{5}{c}{SimpleView \cite{Goyal21simpleview}}
        & \multicolumn{5}{c}{PointNet \cite{charles2017pointnet}}
        \\
        \cmidrule(lr){2-6}\cmidrule(lr){7-11}\cmidrule(lr){12-16}
        & ASR$\uparrow$ & A.Q$\downarrow$ & CD$\downarrow$ & HD$\downarrow$ & A.T$\downarrow$ & ASR$\uparrow$ & A.Q$\downarrow$ & CD$\downarrow$ & HD$\downarrow$ & A.T$\downarrow$ & ASR$\uparrow$ & A.Q$\downarrow$ & CD$\downarrow$ & HD$\downarrow$ & A.T$\downarrow$ 
        \\
        & (\%) & (times) & ($10^{-4}$) & ($10^{-2}$) & (s) & (\%) & (times) & ($10^{-4}$) & ($10^{-2}$) & (s) & (\%) & (times) & ($10^{-4}$) & ($10^{-2}$) & (s) 
        \\
        \midrule
        - & 100.0 & 53.9 & 3.89 & 4.58 & 0.31 & 100.0 & 101.6 & 7.38 & 4.77 & 0.54 & 99.9 & 29.9 & 2.39 & 4.31 & 0.14 
        \\
        \midrule
        Drop (30\%) & 75.4 & 651.6 & 1.68 & 4.12 & 3.26 & 51.1 & 1090.6 & 2.20 & 4.25 & 6.02 & 96.9 & 150.9 & 1.84 & 4.26 & 0.47  
        \\
        Drop (50\%) & 84.4 & 466.1 & 1.21 & 3.71 & 2.18 & 63.7 & 850.0 & 1.49 & 3.77 & 4.94 & 93.6 & 260.1 & 1.70 & 4.20 & 0.78  
        \\
        SOR \cite{zhou2019dupnet} & 49.2 & 1187.6 & 2.45 & 3.34 & 7.82 & 93.3 & 576.0 & 9.73 & 3.77 & 3.96 & 89.7 & 638.9 & 9.11 & 4.20 & 2.33 
        \\
        \bottomrule
    \end{tabular}
    }
    \caption{Resistance of our black-box attack on defended point cloud models. ``Drop (30\%)'' means randomly dropping 30\% of input points.}
    \label{tab:table4}
\end{table*}

\subsection{White-box Performance}

\noindent\textbf{Comparison with Existing Methods.} 
We comprehensively compare our shape-invariant white-box attack with different baselines (\ie, regular gradient-based $l_2$ attack including I-FGM \cite{gu2015towards}, MI-FGM \cite{dong18mifgm} and PGD \cite{Madry18adversarial}; optimization-based point cloud attack including 3d-Adv \cite{Xiang3dadv}, AdvPC \cite{HamdiRTG20advpc} and GeoA$^3$ \cite{geoa3}. 
Specifically, our method are implemented with the step size 0.007 for 50 iterations. 
All adversarial examples are equally constrained by the $l_\infty$-norm ball (radius is 0.16). 
To unify the verification, all baselines adopt the untargeted mode under the same setting. 
We conduct the comparison on a same RTX 3090Ti GPU in terms of attack success rate (ASR), Chamfer distance (CD), Haudorff Distance (HD) and average time cost (A.T). 
The results listed in \Tref{tab:table1} show that our method takes the least geometry distance cost to achieve nearly 100\% ASR and the lower time budget compared to regular gradient-based attacks. 
It is consistent with our intuition of better inheriting the shape invariance from original point clouds during adversarial example generation.

\noindent\textbf{Resistance to Defenses.} 
To verify the attack performance when facing the well-defended point cloud recognition models, we conduct the experiments on defense methods including standard PGD-based adversarial training (AT), point cloud statistical outlier removal SOR \cite{zhou2019dupnet} and DUP-Net \cite{zhou2019dupnet}. 
Trough comparing to three point cloud attack methods in \Tref{tab:table2}, we find that our method outperforms the baselines a lot on outlier removal based defenses, which further proves the proposed shape-invariant perturbations are mainly along the shape surface with fewer outliers.

\begin{figure*}
\centering
\includegraphics[width=1\linewidth]{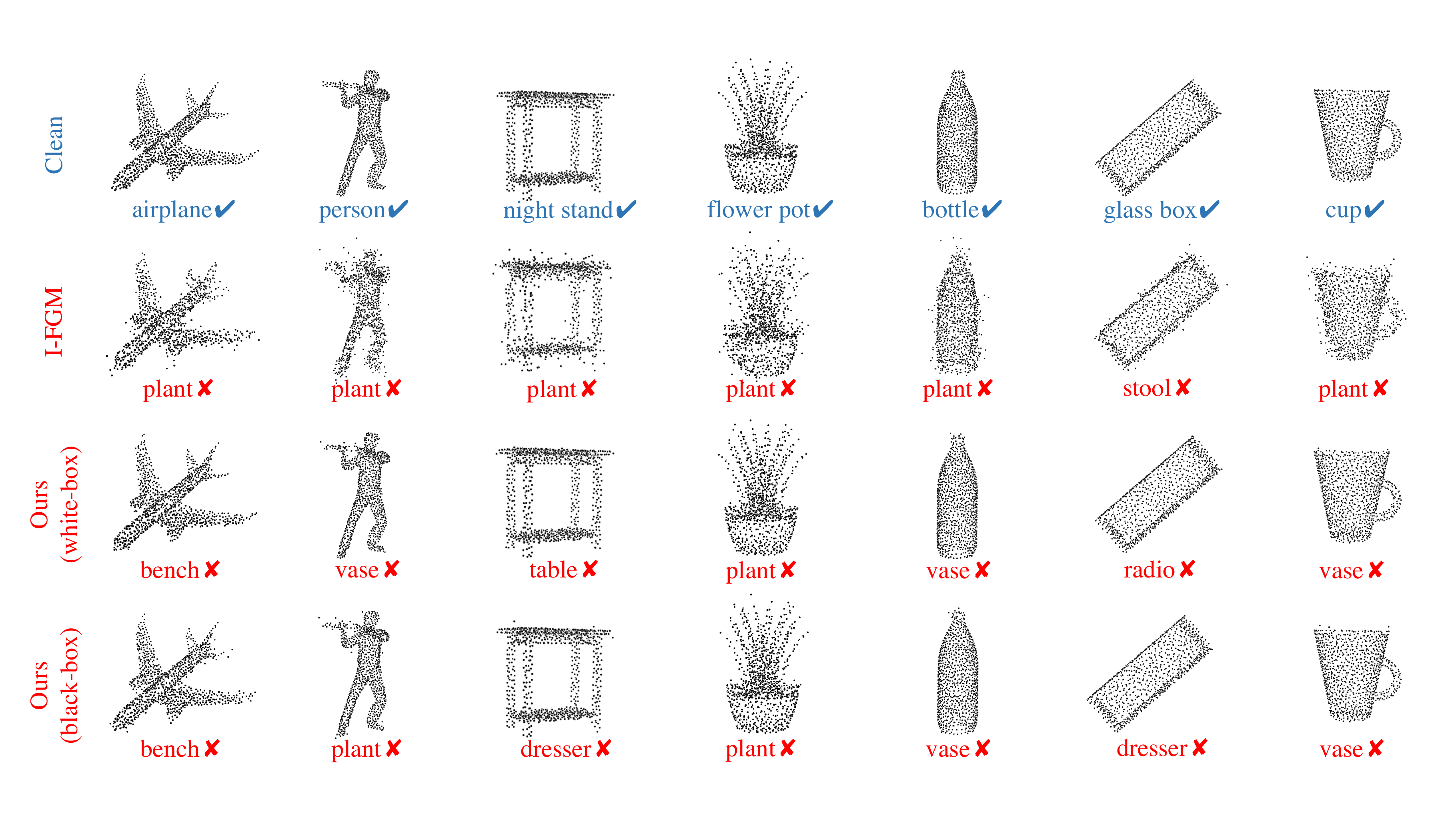}
\vspace{-2em}
\caption{Visualization of clean/adversarial point clouds generated by I-FGM and ours method.}
\label{fig:visual}
\end{figure*}

\subsection{Black-box Performance}

\noindent\textbf{Efficiency on Different Black-box Models.} 
We compare the proposed shape-invariant query-based attack with two related black-box attack algorithms as baselines: SimBA \cite{Guo2019simba} and SimBA+ \cite{YangJHNZ20}. 
SimBA \cite{Guo2019simba} is one of the most famous query-based black-box adversarial attacks that works well in fooling image recognition models, which randomly searches perturbation direction in the whole orthonormal space to perturb each basis. 
Fortunately, we can regard all $N\times 3$ dimensions as the orthonormal bases for SimBA implementation on point clouds. 
SimBA+ \cite{YangJHNZ20} further improve SimBA by narrowing the perturbation searching space with the gradient prior obtained on a pretrained surrogate model. 
But different from image recognition, there are little overlap among pure gradient maps computed on different point cloud models, due to their diverse designs for point feature extraction. 
So the pure gradient prior only brings limited improvement for SimBA++ on unseen point cloud models. 
As results shown in Figure \ref{fig:graph_paconv} and \Tref{tab:table3}, our attack achieves comparable ASR, fewer queries and less point modification than baselines on unseen black-box models. 
In other words, the proposed point-cloud sensitivity map can well depict the internal vulnerability of point cloud recognition models, making the perturbation searching more concrete while maintaining the shape invariance.

\noindent\textbf{Different Step Size $\epsilon$.} 
The attack step size is a crucial and configurable parameter for query-based attack performance. Accordingly, we conduct the evaluation on a variety of step size $\epsilon$ from 0.08 to 0.32. 
In \Fref{fig:graph_paconv}, it can be easily found that the needed queries decreases rapidly under nearly 100\% ASR as the step size gets larger. 
While it significantly strengthens the attack efficacy, the higher $l_2$-norm distance and Chamfer distance indicate more visual quality it has to sacrifice.

\noindent\textbf{Resistance to Defenses.} To investigate the robustness of our shape-invariant black-box attack against different kinds of adversarial defenses, we conduct the experiments on three kinds of defense methods: random point-dropping and input prepocessing based defense SRS \cite{zhou2019dupnet}. 
As shown in \Tref{tab:table4}, for PAConv \cite{Xu21paconv}, SimpleView \cite{Goyal21simpleview} and PointNet \cite{charles2017pointnet}, the ASR on point dropout just degrades up to 36.3\% although the needed queries become more. 
It is worth noted that our method performs even better on PAConv and SimpleView when the dropout defense get enhanced from 30\% to 50 \%. 
Empirically, we attribute it to the larger natural accuracy degradation of these models when dropped points are increasing.
Overall, our black-box shape-invariant attack has the considerable resistance to these defenses.

\subsection{Qualitative Results}

We further visualize the adversarial point clouds and provide the results in \Fref{fig:visual}. 
Since our perturbation is almost along the shape surface, we can hardly find any outliers on the generated adversarial point clouds. 
Notably, our attack is proved to be more threatening with its well-inherited shape invariance and satisfying attack imperceptibility.


\section{Conclusion}

In this paper, we propose Point-Cloud Sensitivity Map by calculating the variation rate of recognition confidence when adding the explicitly constrained shape-invariant perturbation on each point, to reveal the vulnerability of point cloud recognition models when encountering the slight perturbation.
With the guidance of this map, we further propose
Shape-invariant Adversarial Point Clouds to improve the attack imperceptibility and extend the white-box attack to the black-box case. 
To the best of authors' knowledge, it is the first query-based black-box attack on point cloud recognition. 
Experiments show the effectiveness and practicality of our method with its superior shape invariance. 

\section{Acknowledgement} 

This work was supported in part by the Natural Science Foundation of China under Grant U20B2047, 62072421, 62002334, and 62121002, Exploration Fund Project of University of Science and Technology of China under Grant YD3480002001, and by Fundamental Research Funds for the Central Universities under Grant WK2100000011.

\newpage
\large\noindent\textbf{A. The Primary Motivation}
\vspace{0.5em}
\normalsize

As we stated in the above manuscript, the objective of constructing the sensitivity map is to measure the variance of the recognition result when each point encounters the \textbf{imperceptible perturbation}. The reasons for why we use tangent plane gradients are:

\textbf{First}, the saliency map \cite{Zheng2019saliency} is not effective for point perturbing attack we explored. It measures how the recognition result changes when we drop a point, which could be viewed as a \emph{large} step perturb. 
While our perturbing attack relies on a proper measure of the change when we \emph{slightly} perturbed the point.
For example, on query-based black-box attack on PointNet++, saliency map obtains 92.6\% ASR and 987.9 average queries(A.Q), while our method outperform it with 95.8\% ASR and 417.0 A.Q under the same settings. 

\textbf{Second}, 
from the perspective of invisibility, the typical gradient map does not have a perturb direction constrain, so the perturbed points are always out of the surface, which is easy to be detected or removed (Figure \ref{fig:intro_1}(b) of our manuscript). On the contrary, if adding perturbations along the tangent plane, it would be more invisible.

So we realize our tangent gradient map by projecting the gradient map to the tangent planes, which concerns both adversary effectiveness and invisibility.

\vspace{0.5em}
\large\noindent\textbf{B. Limitations and Social Impacts}
\vspace{0.5em}
\normalsize

One of the limitations of shape-invariant adversarial attack is \textit{its unsatisfactory applicability on surfaces with large curvature.} 
On the one hand, in Eq.(1) of our main paper, the normal vector simulation about the investigated point with limited neighbor points might be not entirely accurate, especially when this point is located at the surface with large curvature (\eg, the edges or corners of 3D shapes). 
On the other hand, since we design the perturbation along the tangent plane as the ``explicit constrain'' to maintain our claimed shape invariance, the perturbed point is still possible to escape from the surface if this position has very large curvature. 
This limitation partially breaks our shape-invariant assumption, leading to the imperfect visual quality. 
However, the solution to alleviate this problem is also obvious. 
We can adopt much smaller step size and more attack iterations (\ie, more time budget) to realize attack.

The main social concerns about shape-invariant point-cloud attack might be that \textit{it poses a threat to the security of autonomous driving.} 
The attacker can materialize the shape-invariant adversarial point clouds with 3D printing or combine the query-based attack with LiDAR spoofing attack \cite{SunCCM20,CaoXCZPRCFM19} to fool the point cloud recognition module of self-driving cars, leading to the traffic accidents. 
But the prerequisite of successfully realizing such attacks is that attacker needs to get the model output (\eg, at least the predicted logits) to guide his/her attack loops. 
Therefore, unless pure black-box transfer-based attack, the self-driving can still defend itself by robustness enhancement and data leakage reduction based on technologies like strict encryption. 
From the positive perspective, as elaborated in our main paper (\ie, Intro and Social Impact section), our method \textit{provides a more effective evaluation method} for the adversarial robustness of point cloud recognition models, especially for black-box requirement \textit{since the common protection agreement of trade secrets in current industry.} In other words, it facilitates the development of the researches on improving 3D recognition robustness.

\begin{figure*}
\centering
\includegraphics[width=0.9\linewidth]{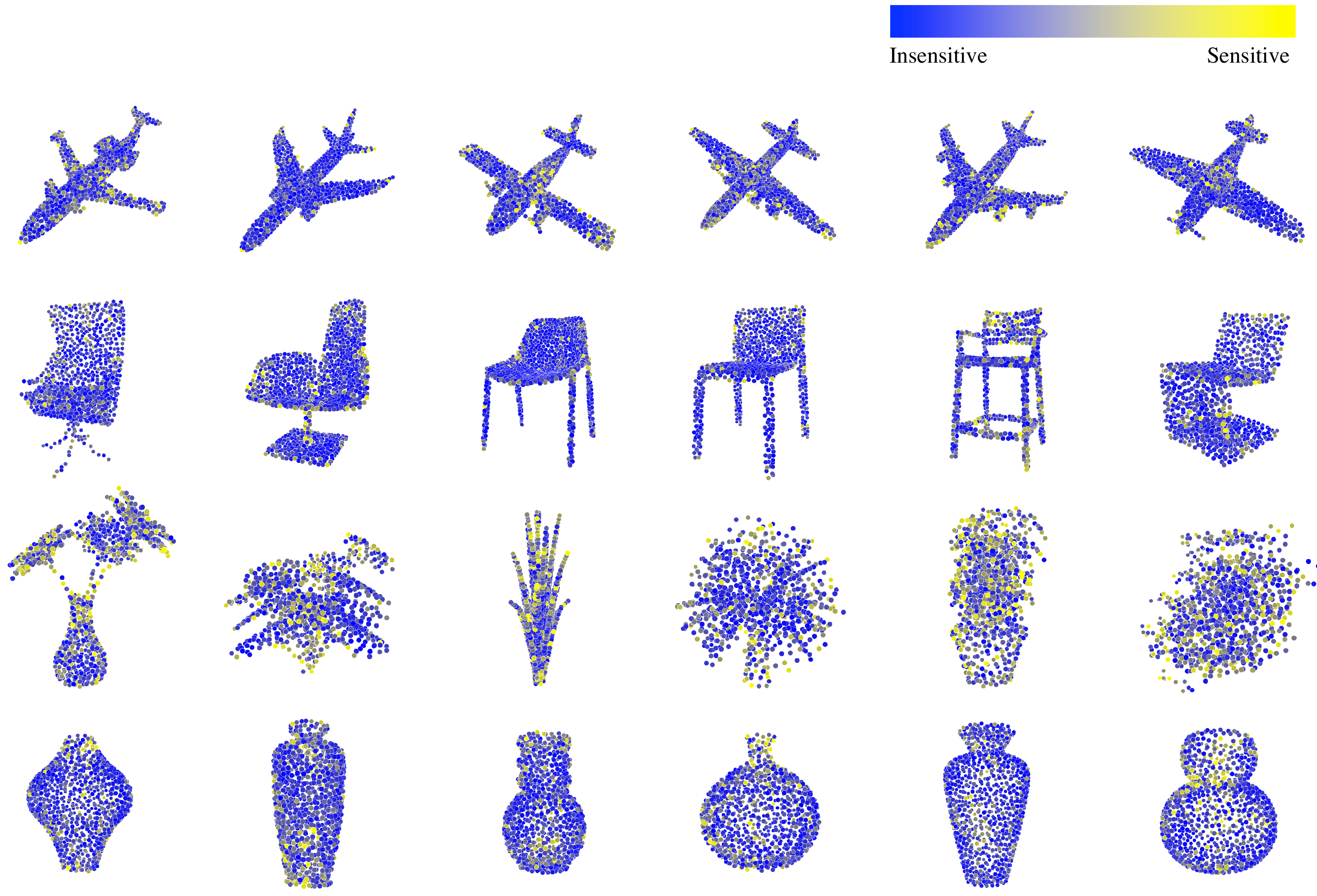}
\caption{Visualization results of the proposed point-cloud sensitivity maps obtained on CurveNet.}
\label{fig:supp_1}
\end{figure*}

\vspace{0.5em}
\large\noindent\textbf{C. Proof of Theorem 1}
\normalsize

\begin{proof}
First, we need to clarify the translation relationship between the original coordinate system origin $\boldsymbol{O}$ and the new coordinate system origin $\boldsymbol{O}'$. 
As illustrated in Figure 3 in our main paper, since $\boldsymbol{O}'$ is the projection of $\boldsymbol{O}$ on the tangent plane $\Omega_i$, so $\overrightarrow{\boldsymbol{O'O}}$ is parallel to the normal vector $\boldsymbol{n}_i$. Thus the translation relationship can be calculated by
\begin{equation}
    \overrightarrow{\boldsymbol{OO'}} = k \boldsymbol{n}_i \Leftrightarrow \boldsymbol{O} (0, 0, 0), \boldsymbol{O}' (k n_{i1}, k n_{i2}, k n_{i3}), 
\end{equation}
where $k = (\boldsymbol{p}_i \cdot \boldsymbol{n}_i)$ is the module length of vector $\boldsymbol{OO'}$.

Then, with the coordinate of $\boldsymbol{O}'$, we can easily obtain the vector $\overrightarrow{\boldsymbol{BA}}$ and $\overrightarrow{\boldsymbol{CO'}}$ that define the directions of the transformed axis $x'$ and $y'$ respectively. 
Based on the position representations in Eq.(3) of our main paper, we have 
\begin{align}
    \overrightarrow{\boldsymbol{BA}} &= (\frac{k}{n_{i1}}, -\frac{k}{n_{i2}}, 0), \\
    \overrightarrow{\boldsymbol{CO'}} &= (kn_{i1}, kn_{i2}, kn_{i3} - \frac{k}{n_{i3}}). 
\end{align}
Therefore, if we define the standard orthogonal coordinate bases of the original coordinate system $\boldsymbol{O}-xyz$ as $\overrightarrow{\boldsymbol{x}}$, $\overrightarrow{\boldsymbol{y}}$ and $\overrightarrow{\boldsymbol{z}}$, the standard orthogonal coordinate bases $\overrightarrow{\boldsymbol{x}}'$, $\overrightarrow{\boldsymbol{y}}'$ and $\overrightarrow{\boldsymbol{z}}'$ of the new coordinate system $\boldsymbol{O'}-x'y'z'$ can be formulated as
\begin{align}
    \overrightarrow{\boldsymbol{x}}' &= \frac{\overrightarrow{\boldsymbol{BA}}}{|\overrightarrow{\boldsymbol{BA}}|}, \quad \overrightarrow{\boldsymbol{y}}' = \frac{\overrightarrow{\boldsymbol{CO'}}}{|\overrightarrow{\boldsymbol{CO'}}|}, \quad
    \overrightarrow{\boldsymbol{z}}' = \frac{\overrightarrow{\boldsymbol{O'O}}}{|\overrightarrow{\boldsymbol{O'O}}|}, \\
    \overrightarrow{\boldsymbol{x}}' &= \Big(\frac{n_{i2}}{\sqrt{1 - n_{i3}^2}}\Big) \overrightarrow{\boldsymbol{x}} + \Big(-\frac{n_{i1}}{\sqrt{1 - n_{i3}^2}}\Big) \overrightarrow{\boldsymbol{y}} + 0 \overrightarrow{\boldsymbol{z}}, \\
    \overrightarrow{\boldsymbol{y}}' &= \Big(\frac{n_{i1}n_{i3}}{\sqrt{1 - n_{i3}^2}}\Big) \overrightarrow{\boldsymbol{x}} + \Big(\frac{n_{i2}n_{i3}}{\sqrt{1 - n_{i3}^2}}\Big) \overrightarrow{\boldsymbol{y}} \\
    &+ \Big(-\sqrt{1 - n_{i3}^2}\Big) \overrightarrow{\boldsymbol{z}}, \\
    \overrightarrow{\boldsymbol{z}}' &= n_{i1} \overrightarrow{\boldsymbol{x}} + n_{i2} \overrightarrow{\boldsymbol{y}} + n_{i3} \overrightarrow{\boldsymbol{z}}. 
\end{align}
The above relationships between the bases of old/new orthogonal coordinate system reveal the rotation transformation $f_{ir}$ of axes. 
Hence we can reorganize the above equations to get the translation transformation matrix, \ie, 
\begin{equation}
    \boldsymbol{R}_i = 
    \left(
    \begin{array}{ccc}
    \frac{n_{i2}}{\sqrt{1-n_{i3}^2}} & \frac{-n_{i1}}{\sqrt{1-n_{i3}^2}} & 0 \\
    \frac{n_{i1}n_{i3}}{\sqrt{1-n_{i3}^2}} & \frac{n_{i2}n_{i3}}{\sqrt{1-n_{i3}^2}} & -\sqrt{1-n_{i3}^2} \\
    n_{i1} & n_{i2} & n_{i3}
    \end{array}
    \right), 
\end{equation}
which denotes the rotation transformation from $\boldsymbol{O}-xyz$ to $\boldsymbol{O'}-x'y'z'$. 
Note that the denominator $\sqrt{1-n_{i3}^2}$ is equal to $0$ when $n_{i3} = 1$. 
Thus we further consider the limit for this boundary case when $n_{i1} = n_{i2} = 0$ and $n_{i3} = 1$, \ie, 
\begin{equation}
    \lim_{|n_{i3}| \rightarrow 1} \boldsymbol{R}_i = 
    \left(
    \begin{array}{ccc}
    1 / \sqrt{2} & -1 / \sqrt{2} & 0 \\
    n_{i3} / \sqrt{2} & n_{i3} / \sqrt{2} & 0 \\
    0 & 0 & n_{i3}
    \end{array}
    \right). 
\end{equation}
Likewise, the translation transformation matrix can be obtained from the origin shift $\overrightarrow{\boldsymbol{OO'}}$ as
\begin{equation}
    \boldsymbol{T}_i = (\boldsymbol{p}_i \cdot \boldsymbol{n}_i) \boldsymbol{n}_i,
\end{equation}
which denotes the translation transformation from $\boldsymbol{O}-xyz$ to $\boldsymbol{O'}-x'y'z'$.
To get the transformed coordinate of $\boldsymbol{p}_i$ (\ie, its coordinate in the new coordinate system $\boldsymbol{O'}-x'y'z'$), we first shift it according to $\overrightarrow{\boldsymbol{OO'}}$ and then rotate it, which can be formulated as
\begin{equation}
    \boldsymbol{p}'_i  = \boldsymbol{R}_i (\boldsymbol{p}_i + \boldsymbol{T}_i ). 
\end{equation}
By contrast, we can first apply rotation and then apply translation to realize the reverse coordinate transformation by
\begin{equation}
    \boldsymbol{p}_i  = \boldsymbol{R}_i^\top \boldsymbol{p}'_i - \boldsymbol{T}_i, 
\end{equation}
When we combine the above two equations, we can get
\begin{equation}
    \boldsymbol{p}'_i  = \boldsymbol{R}_i (\boldsymbol{p}_i + \boldsymbol{T}_i ) = \boldsymbol{R}_i (\boldsymbol{R}_i^\top \boldsymbol{p}'_i) = (\boldsymbol{R}_i \boldsymbol{R}_i^\top) \boldsymbol{p}'_i, 
\end{equation}
where we can get $\boldsymbol{R}_i \boldsymbol{R}_i^\top = \boldsymbol{I}$, In this way, the $l_2$-norm of the rotation matrix $\boldsymbol{R}_i^\top$ can be determined by the maximal eigenvalue denoted by $\sigma_{max}$, \ie, 
\begin{equation}
    \|\boldsymbol{R}_i^\top\|_2^2 = \sigma_{max}(\boldsymbol{R}_i \boldsymbol{R}_i^\top) = 1.
\end{equation}
Thus the proof of Theorem 1 is completed.
\end{proof}

\vspace{0.5em}
\large\noindent\textbf{D. More Visualization Results}
\vspace{0.5em}
\normalsize

We provide more visual results for the proposed point-cloud sensitivity maps in Figure \ref{fig:supp_1}.

\begin{table*}[t]
	\footnotesize
    \centering
    \setlength{\tabcolsep}{1mm}{
    \begin{tabular}{lcp{10mm}<{\centering}p{10mm}<{\centering}p{10mm}<{\centering}p{10mm}<{\centering}p{10mm}<{\centering}p{10mm}<{\centering}p{10mm}<{\centering}p{10mm}<{\centering}p{10mm}<{\centering}p{10mm}<{\centering}p{10mm}<{\centering}p{10mm}<{\centering}}
        \toprule
        \multirow{3}{*}{Attack}
        & \multirow{3}{*}{Defense}
        & \multicolumn{4}{c}{PointNet \cite{charles2017pointnet}} 
        & \multicolumn{4}{c}{DGCNN \cite{dgcnn}}
        & \multicolumn{4}{c}{CurveNet \cite{xiang2021curvenet}}
        \\
        \cmidrule(lr){3-6}\cmidrule(lr){7-10}\cmidrule(lr){11-14}
        & & ASR$\uparrow$ & CD$\downarrow$ & HD$\downarrow$ & A.T$\downarrow$ & ASR$\uparrow$ & CD$\downarrow$ & HD$\downarrow$ & A.T$\downarrow$ & ASR$\uparrow$ & CD$\downarrow$ & HD$\downarrow$ & A.T$\downarrow$ 
        \\
        & & (\%) & ($10^{-4}$) & ($10^{-2}$) & (s) & (\%) & ($10^{-4}$) & ($10^{-2}$) & (s) & (\%) & ($10^{-4}$) & ($10^{-2}$) & (s) 
        \\
        \midrule
        I-FGM \cite{gu2015towards} & - & 99.9 & 6.68 & 3.88 & \textbf{1.05} & 97.8 & 24.46 & 3.60 & 2.12 & 98.2 & 16.20 & 3.77 & 11.68  
        \\
        MI-FGM \cite{dong18mifgm} & - & 97.5 & 21.54 & 4.88 & 1.08 & 60.2 & 114.09 & 5.51 & \textbf{2.11} & 81.1 & 80.92 & 5.42 & \textbf{11.56}   
        \\
        PGD \cite{Madry18adversarial} & - & 99.9 & 6.63 & 3.88 & 1.06 & 97.8 & 24.26 & 3.61 & 3.16 & 98.2 & 16.33 & 3.77 & 11.58   
        \\
        \midrule
        3d-Adv \cite{Xiang3dadv} & - & \textbf{100.0} & 7.30 & 3.75 & 5.09 & \textbf{100.0} & 15.43 & 5.13 & 19.10 & \textbf{100.0} & 15.47 & 3.59 & 121.94   
        \\
        AdvPC \cite{HamdiRTG20advpc} & - & 83.6 & 10.98 & 4.43 & 2.73 & 95.4 & 13.42 & 3.24 & 7.83 & 68.6 & 16.17 & 3.74 & 56.64  
        \\
        \textbf{Ours} & - & 95.2 & \textbf{3.59} & \textbf{3.46} & 1.26 & 93.5 & \textbf{9.08} & \textbf{2.95} & 3.69 & 90.9 & \textbf{8.29} & \textbf{3.58} & 21.41   
        \\
        \bottomrule
    \end{tabular}
    }
    \caption{Quantitative comparison on ShapeNetPart between our white-box shape-invariant attack and different white-box attacks, on attack success rate (ASR), Chamfer distance (CD), Hausdorff distance (HD) and average time budget for each adversarial point cloud generation (A.T), where Chamfer distance is multiplied by $10^4$ and Hausdorff distance is multiplied by $10^2$ for better comparison.}
    \label{tab:table1}
\end{table*}

\begin{table*}[h!]
	\footnotesize
    \centering
    \setlength{\tabcolsep}{1mm}{
    \begin{tabular}{lp{8.5mm}<{\centering}p{8.5mm}<{\centering}p{8.5mm}<{\centering}p{8.5mm}<{\centering}p{8.5mm}<{\centering}p{8.5mm}<{\centering}p{8.5mm}<{\centering}p{8.5mm}<{\centering}p{8.5mm}<{\centering}p{8.5mm}<{\centering}p{8.5mm}<{\centering}p{8.5mm}<{\centering}p{8.5mm}<{\centering}p{8.5mm}<{\centering}p{8.5mm}<{\centering}}
        \toprule
        \multirow{3}{*}{Attack}
        & \multicolumn{5}{c}{SimpleView \cite{Goyal21simpleview}} 
        & \multicolumn{5}{c}{PAConv \cite{Xu21paconv}}
        & \multicolumn{5}{c}{CurveNet \cite{xiang2021curvenet}}
        \\
        \cmidrule(lr){2-6}\cmidrule(lr){7-11}\cmidrule(lr){12-16}
        & ASR$\uparrow$ & A.Q$\downarrow$ & CD$\downarrow$ & HD$\downarrow$ & A.T$\downarrow$ & ASR$\uparrow$ & A.Q$\downarrow$ & CD$\downarrow$ & HD$\downarrow$ & A.T$\downarrow$ & ASR$\uparrow$ & A.Q$\downarrow$ & CD$\downarrow$ & HD$\downarrow$ & A.T$\downarrow$ 
        \\
        & (\%) & (times) & ($10^{-4}$) & ($10^{-2}$) & (s) & (\%) & (times) & ($10^{-4}$) & ($10^{-2}$) & (s) & (\%) & (times) & ($10^{-4}$) & ($10^{-2}$) & (s) 
        \\
        \midrule
        SimBA \cite{Guo2019simba} & 100.0 & 119.5 & 8.65 & 5.35 & 0.61 & 100.0 & 73.2 & 4.51 & 4.93 & 0.36 & 100.0 & 114.5 & 4.67 & 4.93 & 13.51
        \\
        SimBA+ \cite{YangJHNZ20} & 100.0 & 115.8 & 10.01 & 11.64 & 0.69 & 100.0 & 67.4 & 5.02 & 9.58 & 0.36 & \textbf{100.0} & 98.9 & 5.29 & 9.76 & 12.94 
        \\
        \textbf{Ours-D} & \textbf{100.0} & \textbf{101.6} & \textbf{7.38} & \textbf{4.77} & \textbf{0.54} & \textbf{100.0} & \textbf{53.9} & \textbf{3.89} & \textbf{4.58} & 0.31 & 99.9 & \textbf{81.5} & \textbf{3.97} & \textbf{4.59} & \textbf{8.77} 
        \\
        \textbf{Ours-P} & 98.6 & 105.7 & 8.03 & 4.88 & 0.57 & 99.7 & 55.3 & 4.60 & 4.77 & \textbf{0.30} & 98.9 & 86.6 & 4.51 & 4.78 & 9.36 
        \\
        \bottomrule
    \end{tabular}
    }
    \caption{Quantitative comparison on ModelNet40 between our shape-invariant black-box attack and different black-box attacks with step size 0.32, on attack success rate (ASR), average query cost (A.Q), Chamfer distance (CD), Hausdorff distance (HD) and average time budget (A.T), where ``Ours-D'' means choosing DGCNN as surrogate model $\mathcal{H}_w$ and ``Ours-P'' means choosing PointNet as $\mathcal{H}_w$. }
    \label{tab:table2}
\end{table*}

\vspace{0.5em}
\large\noindent\textbf{E. More Experimental Results}
\vspace{0.5em}
\normalsize

\noindent\textbf{(1). White-box Performance on ShapeNetPart} 

Except for ModelNet40 \cite{Wu2015modelnet}, we also compare the propsoed shape-invaraint white-box attack with baselines on ShapeNetPart \cite{Yi16shapenet}. 
We train three popular point cloud models (\ie, PointNet \cite{charles2017pointnet}, PointNet++ \cite{charles2017pointnet++} and CurveNet \cite{xiang2021curvenet}) on ShapeNetPart for 150 epochs. 
Initially, all of their clean recognition accuracy is nearly 99\%. 
Similarly, we adopt the same attack settings with the settings clarified in our main paper (Sec 4.3). 
As the results listed in Table \ref{tab:table1}, our method can still achieve the low Chamfer distance \cite{Fan2017chamfer} while maintaining over 90\% attack success rate (ASR). 
As an gradient-based iterative attack method, it is hard-won for our method to achieve high ASR (effectiveness), low geometry distances (invisibility) and low A.T (efficiency) at the same time.

\noindent\textbf{(2). Black-box Performance with PointNet as $\mathcal{H}_w$} 

Since the results reported in the main paper are obtained by utilizing DGCNN \cite{dgcnn} as the surrogate model, the readers may be just wondering that \textit{what if using the weaker model (\eg, PointNet \cite{charles2017pointnet}) as the surrogate model?} 
As showcased in Table \ref{tab:table2}, even when we take PointNet as the weak surrogate model to implement our query-based attack, the attack performance just has few degradation. 
Specifically, when attacking on three of the most advanced point cloud recognition models including SimpleView \cite{Goyal21simpleview}, PAConv \cite{Xu21paconv} and CurveNet \cite{xiang2021curvenet}, our performance on query cost and visual quality are still much better than SimBA \cite{Guo2019simba} and SimBA+ \cite{YangJHNZ20} though few ASR is sacrificed.

\clearpage
{\small
\bibliographystyle{ieee_fullname}
\bibliography{egbib}
}

\end{document}